\documentclass[10pt,twocolumn,letterpaper]{article}

\usepackage{iccv}
\usepackage{times}
\usepackage{epsfig}
\usepackage{graphicx}
\usepackage{amsmath}
\usepackage{amssymb}
\usepackage{booktabs}
\usepackage{url}
\usepackage{amsfonts}       
\usepackage{nicefrac}       
\usepackage{microtype}      
\usepackage{subfig}
\usepackage{algorithm}
\usepackage{algorithmic}
\usepackage{wrapfig}
\usepackage{epsfig}
\usepackage{multirow}
\usepackage[dvipsnames]{xcolor}
\usepackage{arydshln}
\usepackage{threeparttable}
\usepackage{colortbl}
\usepackage{enumitem}
\usepackage{siunitx}
\usepackage{bm}
\usepackage[export]{adjustbox}
\usepackage{listings} 
\usepackage{pifont}
\usepackage{dblfloatfix}
\usepackage[numbers,sort&compress]{natbib}

\definecolor{delay}{RGB}{230,94,42}
\definecolor{mywarning}{RGB}{233,144,61}
\definecolor{mygray}{gray}{.9}
\definecolor{ggray}{RGB}{127,127,127}
\definecolor{reda}{RGB}{192,0,0}
\definecolor{redb}{RGB}{217,148,143}
\definecolor{myyellow}{RGB}{190,144,0}
\definecolor{mygreen}{RGB}{80,100,40}
\definecolor{myblue}{RGB}{30,90,100}

\newcommand{\reshl}[2]{
\textbf{#1}\fontsize{7.5pt}{1em}\selectfont{\color{mygreen}{~$\uparrow$\textbf{#2}}}
}

\newcommand{\delay}[2]{
#1\fontsize{7.5pt}{1em}\selectfont{\color{delay}{~\textbf{#2}}}
}

\newcommand{\cmark}{\ding{51}}%
\makeatletter
\newcommand{\thickhline}{
  \noalign {\ifnum 0=`}\fi \hrule height 1pt
  \futurelet \reserved@a \@xhline
}

\usepackage{tabulary}
\newcolumntype{x}[1]{>{\centering\arraybackslash}p{#1pt}}
\newcolumntype{y}[1]{>{\raggedright\arraybackslash}p{#1pt}}
\newcolumntype{z}[1]{>{\raggedleft\arraybackslash}p{#1pt}}

\usepackage{algorithm}
\usepackage{listings}

\usepackage{etoolbox}
\makeatletter
\AfterEndEnvironment{algorithm}{\let\@algcomment\relax}
\AtEndEnvironment{algorithm}{\kern2pt\hrule\relax\vskip3pt\@algcomment}
\let\@algcomment\relax
\newcommand\algcomment[1]{\def\@algcomment{\footnotesize#1}}
\renewcommand\fs@ruled{\def\@fs@cfont{\bfseries}\let\@fs@capt\floatc@ruled
  \def\@fs@pre{\hrule height.8pt depth0pt \kern2pt}%
  \def\@fs@post{}%
  \def\@fs@mid{\kern2pt\hrule\kern2pt}%
  \let\@fs@iftopcapt\iftrue}
\makeatother

\usepackage{mathtools, nccmath}
\usepackage{comment}
\usepackage{xspace}
\usepackage{ragged2e}
\newcolumntype{P}[1]{>{\RaggedRight\hspace{0pt}}p{#1}}
\newcolumntype{X}[1]{>{\RaggedRight\hspace*{0pt}}p{#1}}
\usepackage{tcolorbox}
\usepackage[edges]{forest}
\colorlet{linecol}{black!75}
\usepackage{xkcdcolors} 

\colorlet{mhpurple}{Plum!80}

\definecolor{mygray2}{gray}{.6}
\definecolor{mygray3}{gray}{.3}
\definecolor{mygray}{gray}{.9}
\definecolor{mywarning}{RGB}{233,144,61}
\definecolor{ggray}{RGB}{127,127,127}
\definecolor{reda}{RGB}{192,0,0}
\definecolor{redb}{RGB}{217,148,143}
\definecolor{myyellow}{RGB}{190,144,0}
\definecolor{mygreen}{RGB}{0,153,0}
\definecolor{mygreen2}{RGB}{153,255,153}
\definecolor{myred}{RGB}{177,35,15}
\definecolor{myblue}{RGB}{58,79,116}
\definecolor{myy}{RGB}{254,203,50}
\definecolor{myy2}{RGB}{191,144,0}
\definecolor{myg}{RGB}{205,205,205}
\definecolor{myg2}{RGB}{80,80,80}
\definecolor{codegreen}{RGB}{79,126,127}
\definecolor{codedefine}{RGB}{153,54,159}
\definecolor{codefunc}{RGB}{73,122,234}
\definecolor{codecall}{RGB}{73,122,234}
\definecolor{codepro}{RGB}{212,96,80}
\definecolor{codedim}{RGB}{89,152,195}

\usepackage[pagebackref=true,breaklinks=true,letterpaper=true,colorlinks,bookmarks=false]{hyperref}
\usepackage{amsthm}
\theoremstyle{definition}
\newtheorem{definition}{Definition}[subsection]

\usepackage[capitalize]{cleveref}
\crefname{section}{Sec.}{Secs.}
\Crefname{section}{Section}{Sections}
\Crefname{table}{Table}{Tables}
\crefname{table}{Tab.}{Tabs.}

\DeclareMathOperator*{\argmax}{argmax}

\iccvfinalcopy 

\def\iccvPaperID{5876} 

\ificcvfinal\fi

\begin{document}

\title{$_{\!\!\!\!\!}$\textsc{LogicSeg}:$_{\!}$~Parsing$_{\!}$~Visual Semantics with Neural Logic Learning and Reasoning$_{\!\!\!\!\!\!\!\!\!\!\!\!}$}

\author{
Liulei Li\textsuperscript{1,~2}, Wenguan Wang\textsuperscript{1}\footnotemark[1],~~Yi Yang\textsuperscript{1}\\
\small \textsuperscript{1} ReLER, CCAI, Zhejiang University~~\textsuperscript{2} ReLER, AAII, University of Technology Sydney\\
\small\url{https://github.com/lingorX/LogicSeg/}
}

\maketitle
\ificcvfinal\fi

\begin{abstract}
\footnotetext[1]{Corresponding author: Wenguan Wang.}
Current high-performance semantic segmentation models are$_{\!}$ purely$_{\!}$ data-driven$_{\!}$ sub-symbolic$_{\!}$ approaches$_{\!}$ and$_{\!}$ blind$_{\!}$ to$_{\!}$ the$_{\!}$ structured$_{\!}$ nature$_{\!}$ of$_{\!}$ the$_{\!}$ visual$_{\!}$ world.$_{\!}$ This$_{\!}$ is in$_{\!}$  stark$_{\!}$  contrast$_{\!}$  to$_{\!}$ human$_{\!}$  cognition$_{\!}$  which$_{\!}$ abstracts$_{\!}$  visual$_{\!}$ percep- tions$_{\!}$ at$_{\!}$ multiple$_{\!}$ levels$_{\!}$ and$_{\!}$ conducts$_{\!}$ symbolic$_{\!}$ reasoning$_{\!}$ with such structured abstraction. To fill these fundamental gaps, we devise \textsc{LogicSeg}, a holistic visual semantic parser that integrates$_{\!}$  neural$_{\!}$  inductive$_{\!}$  learning$_{\!}$  and$_{\!}$  logic$_{\!}$  reasoning$_{\!}$  with both$_{\!}$  rich$_{\!}$  data$_{\!}$  and$_{\!}$  symbolic$_{\!}$  knowledge.$_{\!}$  In$_{\!}$  particular, the semantic concepts of interest are structured~as~a hierarchy,$_{\!}$ from$_{\!}$ which$_{\!}$ a$_{\!}$ set$_{\!}$ of$_{\!}$ {constraints}$_{\!}$ are$_{\!}$ derived for describing the symbolic relations and {formalized} as first-order logic rules. After fuzzy logic-based continuous {relaxation},~logical formulae are grounded onto data and neural  computational
graphs, hence enabling logic-induced network training. During inference, logical constraints are packaged into an iterative process and injected into the network in a form of {several} matrix multiplications, so as to achieve hierarchy-coherent prediction$_{\!}$ with$_{\!}$ logic$_{\!}$ reasoning.$_{\!}$ These$_{\!}$ designs$_{\!}$ together$_{\!}$ make \textsc{LogicSeg} a general and compact neural-logic machine that is readily integrated into existing segmentation models.\\
\noindent Extensive experiments over four datasets with various segmentation models and backbones verify the effectiveness and generality$_{\!}$ of$_{\!}$ \textsc{LogicSeg}.$_{\!\!}$ We$_{\!}$ believe$_{\!}$ this$_{\!}$ study$_{\!}$ opens$_{\!}$ a$_{\!}$ new avenue$_{\!}$ for$_{\!}$ visual$_{\!}$ semantic$_{\!}$ parsing.
\end{abstract}

\vspace{-10pt}
\section{Introduction}
\vspace{-3pt}
Interpreting$_{\!}$ high-level$_{\!}$ semantic$_{\!}$ concepts of visual stimuli is an integral aspect of human perception and cognition, and has been a subject of interest in computer vision for nearly as$_{\!}$ long$_{\!}$ as$_{\!}$ this$_{\!}$ discipline$_{\!}$ has$_{\!}$ existed.$_{\!}$ As$_{\!}$ an$_{\!}$ exemplar$_{\!}$ task$_{\!}$ of$_{\!}$ vi- sual~semantic interpretation, \textit{semantic segmentation} aims to group pixels into different semantic units. Progress in this field has been notable since the seminal work of fully convolution networks (FCNs)\!~\cite{long2015fully} and been
further advanced by the recent launch  of fully attention networks (Transformer) \cite{vaswani2017attention}.

\begin{figure}[t]
  \vspace{-5px}
  \begin{center}
    \includegraphics[width=\linewidth]{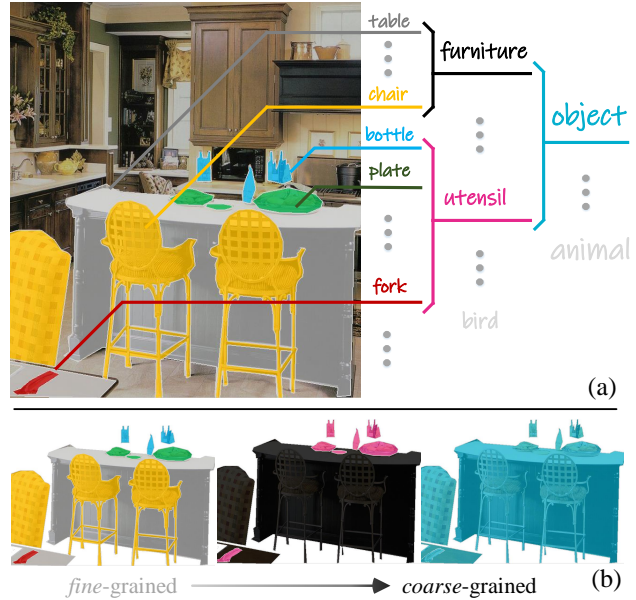}
  \end{center}
   \vspace{-15px}
  \caption{\small{(a) We humans abstract our perception in a structured  manner, and conduct reasoning through symbol manipulation over such multi-level abstraction. (b) We aim to \textit{holistically} interpret visual semantics, through the integration of both data-driven sub-symbolic learning and symbolic knowledge-based logic reasoning.}}
  \label{fig:1}
  \vspace{-10px}
\end{figure}

Despite these technological strides, we still observe current prevalent segmentation systems lack in-depth reflection on some intrinsic nature of human cognition. \textbf{First}, standard segmentation systems simply assume
the semantic concepts in the set of {interest} have no underlying relation and predict all these concepts \textit{exclusively}.$_{\!}$ By contrast, humans interpret a scene by components. For example in Fig.~\ref{fig:1}, we can effortlessly recognize many pieces of \texttt{furniture}, such as \texttt{chair}s and \texttt{table}s, and identify various \texttt{utensil}s, \eg, \texttt{bottle}s, and \texttt{plate}s. Such capacity {of} structured understanding of visual semantics is an innate aspect of human perception~\cite{bill2020hierarchical}, complies with our way of the organization of knowledge~\cite{frankland2020concepts,yang2021multiple}, and has a close relation to many meta-cognitive skills including \textit{compositional generalization} (\ie, making infinite use of finite means)~\cite{janssen2012compositionality}, \textit{systematicity} (\ie, cognitive capacity comes in groups of related behaviours)~\cite{fodor1988connectionism}, and \textit{interpretability} (\ie, interpreting complex concepts with simpler ones)~\cite{pagin2010compositionality,smolensky2022neurocompositionala}. 
Despite its significance and ubiquity,  surprisingly little has been done on the computational modeling of structured visual perception in the segmentation literature.$_{\!}$ Though$_{\!}$ exceptions$_{\!}$ exist$_{\!}$~\cite{xiao2018unified,liang2018dynamic,li2020deep,li2022deep,wang2021hierarchical},$_{\!}$~in$_{\!}$~general$_{\!}$~they
are scattered, lacking systematic study. \textbf{Second}, the latest semantic segmentation systems, {label structure aware or not}, have developed a pure sub-symbolic learning approach.$_{\!}$~They enjoy the advantages of robust distributed representation of concept entities, but struggle with explicit reasoning with the relations among entities by discrete symbolic representations~\cite{zhang2021neural}. Nevertheless, studies in cognition suggest that our perception works at multiple levels of semantic abstraction~\cite{reed2016taxonomic}, intertwined with logical reasoning through manipulation of symbolic knowledge/concepts\!~\cite{kahneman2011thinking}. For example, after recognizing many \texttt{utensil}s from Fig.\!~\ref{fig:1}, we know the scene is more likely a \texttt{kitchen}, rather than a \texttt{bathroom} or \texttt{gym}. This judgement comes as a result of reasoning with some abstract knowledge, such as ``\textit{utensils typically appear}\\
\noindent \textit{in the {kitchen}}'' and ``\textit{utensils are seldom seen in the bathroom},'' which are generalized from our daily experience. The judgement of the scene type may become a belief and in turn cause reallocation of our visual attention~\cite{katsuki2014bottom}, hence driving us to find out more relevant details, such as small \texttt{fork}s.

Filling the gaps identified above calls for a fundamental paradigm shift: \textbf{i)} moving away from  pixel-wise `flat' classification towards semantic structure-aware parsing; and \textbf{ii)} moving away from the extreme of pure distributed representation learning towards an ambitious hybrid which {combines} both powerful sub-symbolic learning and principled symbolic reasoning. To embrace this {change}, we develop \textsc{LogicSeg}, a structured visual parser which exploits
neural computing and symbolic logic in a neural-symbolic framework for holistic$_{\!}$ visual semantic learning and reasoning.$_{\!}$ In particular, given$_{\!}$ a$_{\!}$ set$_{\!}$ of$_{\!}$ hierarchically-organized$_{\!}$ semantic$_{\!}$ concepts as~background knowledge and parsing target, we first use \textit{first-order logic}, a powerful declarative language, to comprehensively specify relations among semantic classes. After \textit{fuzzy logic} based relaxation, the logical formulae~of~hi- erarchy$_{\!}$ constraints$_{\!}$ can$_{\!}$ be$_{\!}$ grounded$_{\!}$ on$_{\!}$ data.$_{\!}$ During$_{\!}$ training,  each logical constraint is converted into a differentiable loss function$_{\!}$ for$_{\!}$ gradient$_{\!}$ descent$_{\!}$ optimization.$_{\!}$ During$_{\!}$ inference,$_{\!}$\\
\noindent the$_{\!}$ logical$_{\!}$ constraints$_{\!}$ are$_{\!}$ involved$_{\!}$ into$_{\!}$ an$_{\!}$ iterative$_{\!}$ process,$_{\!}$ and$_{\!}$ calculated$_{\!}$ in$_{\!}$ matrix$_{\!}$ form.$_{\!}$ This$_{\!}$ not only$_{\!}$ ensures$_{\!}$  the$_{\!}$ obser- vance$_{\!}$ of$_{\!}$ the$_{\!}$ compositional$_{\!}$ semantic$_{\!}$ structure$_{\!}$ but$_{\!}$ also$_{\!}$ binds logic reasoning into  network$_{\!}$ feed-forward$_{\!}$ prediction.

By accommodating logic-based symbolic rules into network training and inference, our \textsc{LogicSeg} \textbf{i)} blends statistical learning with symbolic reasoning, \textbf{ii)} obtains better performance, and \textbf{iii)} guarantees its parsing behavior compliant with the logically specified symbolic knowledge. We also remark that our study is relevant to a field of research called \textit{neural-symbolic computing} (NSC)$_{\!}$~\cite{garcez2019neural,kautz2022third,wang2022towards}. With~the promise of integrating two critical cognitive abilities \cite{valiant2003three}: inductive learning (\ie, the ability to learn general principles from experience) and deductive reasoning (\ie, the ability to draw logical conclusions from what has been learned), NSC has long been a multi-disciplinary research focus and shown superiority in certain application scenarios, such as program generation \cite{valkov2018houdini,nye2020learning,parisotto2017neuro}, and question answering \cite{vedantam2019probabilistic,yi2018neural}. This work unlocks the potential of NSC in visual semantic parsing -- a fundamental, challenging, and large-scale vision task.

\textsc{LogicSeg}$_{\!}$ is$_{\!}$ a$_{\!}$ principled$_{\!}$ framework.$_{\!}$ It$_{\!}$~is$_{\!}$~fully$_{\!}$ compatible with existing segmentation network architectures, with only {minor} modification to~the~classification head and a plug-and-play$_{\!}$ logic-induced$_{\!}$ inference$_{\!}$ module.$_{\!}$
We$_{\!}$ perform$_{\!}$ experi- ments$_{\!}$ on$_{\!}$ four$_{\!}$ datasets$_{\!}$ covering$_{\!}$
wide$_{\!}$ application$_{\!}$ scenarios, including$_{\!}$ automated-driving$_{\!}$ (Mapillary$_{\!\!}$ Vistas$_{\!}$ 2.0$_{\!}$~\cite{neuhold2017mapillary},$_{\!}$ City- scapes$_{\!}$ \cite{cordts2016cityscapes}),$_{\!}$ object-centric$_{\!}$ (Pascal-Part\!~\cite{chen2014detect}),$_{\!}$ and$_{\!}$ daily$_{\!}$ (ADE-20K\!~\cite{zhou2017scene})$_{\!}$ scenes.$_{\!}$ Experimental$_{\!}$ results$_{\!}$ show$_{\!}$ that,$_{\!}$~on$_{\!}$~the$_{\!}$~top$_{\!}$~of various$_{\!}$ segmentation$_{\!}$ models$_{\!}$ (\ie,$_{\!}$ DeepLabV3+$_{\!}$~\cite{chen2018encoder},$_{\!}$ Mask- 2Former$_{\!}$~\cite{cheng2022masked})~and backbones (\ie, ResNet-101$_{\!}$~\cite{he2016deep}, Swin-T$_{\!}$~\cite{liu2021swin}), \textsc{LogicSeg} yields solid performance gains (\textbf{1.12}\%-\textbf{3.29}\%$_{\!}$ mIoU)$_{\!}$ and$_{\!}$ suppresses$_{\!}$ prior$_{\!}$ structured$_{\!}$ alternatives.$_{\!}$ The strong generalization and promising performance of \textsc{LogicSeg} evidence the great potential of integrating symbolic reasoning and sub-symbolic learning in machine perception.

\vspace{-3pt}
\section{Related Work}
\noindent\textbf{Semantic Segmentation.} Since the proposal of fully~convolutional networks (FCNs)~\cite{long2015fully}, research studies in pixel-level semantic interpretation have witnessed {a phenomenal growth}.$_{\!}$ Tremendous$_{\!}$ progress$_{\!}$ has$_{\!}$ been$_{\!}$ achieved$_{\!}$ by,$_{\!}$ for$_{\!}$ exam- ple,$_{\!}$ polishing$_{\!}$ context$_{\!}$ cues$_{\!}$~\cite{ronneberger2015u,yu2015multi,badrinarayanan2017segnet,liu2017deep,lin2017refinenet,zhang2018context,he2019adaptive,hu2020class,yuan2020object,yu2020context,liu2021exploit,Hsiao_2021_ICCV,Jin_2021_ICCV2,Jin_2021_ICCV,li2020content,li2023adversarially,zhou2021group}, investigating boundary information$_{\!}$~\cite{chen2016semantic,ding2019boundary,bertasius2016semantic,li2020improving,yuan2020segfix}, incorporating$_{\!}$  neural$_{\!}$ attention \cite{harley2017segmentation,wang2018non,li2018pyramid,zhao2018psanet,hu2018squeeze,he2019dynamic,li2019expectation,fu2019dual,huang2019ccnet,yang2021collaborative,yang2021associating,yang2022decoupling,han2023html}, adopting data structure-aware learning$_{\!}$~\cite{wang2021exploring,zhou2022rethinking,lianggmmseg,zhang2023boosting,li2023unified}, 
and automating network engineering$_{\!}$~\cite{chen2019fasterseg,liu2019auto,nekrasov2019fast,li2020learning}. More recently, the~engage- ment of advanced Transformer~\cite{vaswani2017attention} architecture, which spe-
 cializes$_{\!}$ in$_{\!}$ long-range$_{\!}$ dependency$_{\!}$ modeling,$_{\!}$ is$_{\!}$ widely$_{\!}$ viewed$_{\!}$ as a promising route for further development$_{\!}$~\cite{xie2021segformer,strudel2021segmenter,zheng2021rethinking,cheng2021maskformer,cheng2022masked,liang2023clustseg,feng2023clustering}.

Though impressive, existing segmentation solutions are mainly aware of straightforward prediction for \textit{flatten} labels. They are largely blind to the rich structures among semantic concepts and lack an explicit mechanism for symbol manipulation/logical calculus, which is what
distinguishes humans from other animals~\cite{deacon1997co,russell2016artificial,horst2005computational}. This work represents a small yet solid step towards addressing these fundamental limitations through an integrated neural-logic machine, and inspects semantic segmentation from a brand-new standpoint.

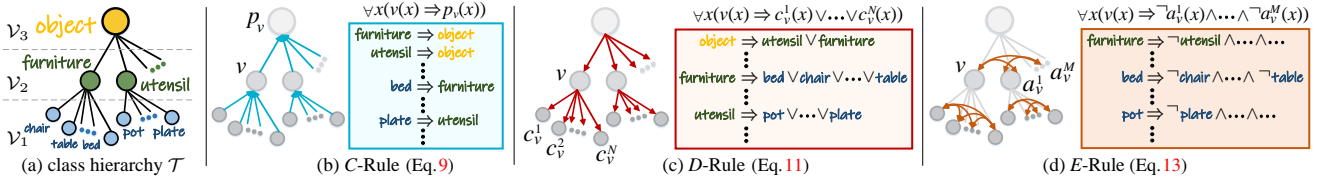
\begin{figure*}[t]
   \vspace{-10pt}
   \begin{center}
      \includegraphics[width=\linewidth]{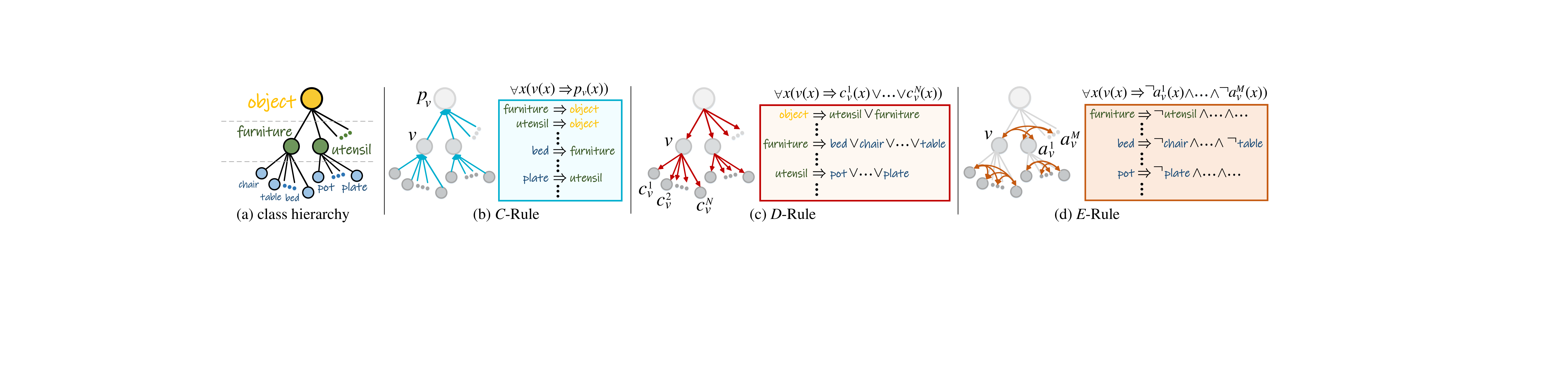}
    \put(-344.1,3.22){\scriptsize (Eq.\!~\ref{eq:CL1})}
    \put(-213.3,3.22){\scriptsize (Eq.\!~\ref{eq:DL1})}
    \put(-70.3,3.22){\scriptsize (Eq.\!~\ref{eq:EL1})}
    \put(-434.3,2.90){\scriptsize $\mathcal{T}$}
    \put(-495.1,13.4){\scriptsize $\mathcal{V}_1$}
    \put(-495.3,34.0){\scriptsize $\mathcal{V}_2$}
    \put(-495.3,53.8){\scriptsize $\mathcal{V}_3$}
   \end{center}
   \vspace{-18pt}
   \captionsetup{font=small}
   \caption{\small{Illustration of the (a) class hierarchy $\mathcal{T}$, and (b-d) abstract relational knowledge specified by first-order logic formulae (\S\ref{sec:knolwedge}).}}
   \vspace{-12pt}
   \label{fig:2}
\end{figure*}

\noindent\textbf{Label Structure-aware Semantic Segmentation.} Till now, only a rather small number of deep learning based segmentation models$_{\!}$~\cite{xiao2018unified,wang2019learning,wang2020hierarchical,ke2022unsupervised,li2022deep} are built with structured label taxonomies. The origin of this line of research can be traced back to the task of \textit{image parsing}$_{\!}$~\cite{wang2019learning,wang2020hierarchical,tu2005image,sudderth2005learning,sudderth2008describing,han2008bottom,yao2012describing} raised in the pre-deep learning era. Basically, image parsing seeks for a holistic explanation of visual observation:$_{\!}$ scenes$_{\!}$ can$_{\!}$~be
understood as a sum of novel objects, and the objects can be further broken down into fine-grained parts. In the deep learning era, the majority of structured segmentation models are dedicated to \textit{human parsing}$_{\!}$~\cite{zhu2018progressive,wang2019learning,wang2020hierarchical,ji2020learning}, which is customized to human-part~relation understanding. As for the case of general-purpose segmentation, there {are} far rare  lite-
rature$_{\!}$~\cite{xiao2018unified,liang2018dynamic,li2020deep,ke2022unsupervised,li2022deep}, and many of them incorporate label taxonomies into the network topology, losing generality$_{\!}$~\cite{xiao2018unified,liang2018dynamic,li2020deep}. As a notable exception, \cite{li2022deep} converts the task as \textit{pixel-wise multi-label classification} and exploits the class hierarchy for training regularization, with only trivial architectural change.

In a nutshell, previous efforts highlight the limits of standard segmentation models for semantic structures. However, they typically \textbf{i)} resolve to stand on the side of sub-symbolic learning,$_{\!}$ \textbf{ii)}$_{\!}$  make$_{\!}$ usage$_{\!}$ of$_{\!}$ only$_{\!}$ fragments$_{\!}$ of$_{\!}$ structured$_{\!}$ relations (for instance, the exclusion relation is neglected by \cite{li2022deep}), \textbf{iii)} lack structure-ware inference, and/or \textbf{iv)} rely on sophisticated and specialized neural structures. By contrast, we formulate the structured task into a neural-symbolic framework. We derive a comprehensive set of symbolic relational knowledge in the form of first-order logic and deeply embed logical constraints into network training and inference. Our algorithm is a general framework that is applicable to existing standard hierarchy-agnostic segmentation architectures.

\noindent\textbf{Neuro-Symbolic Computing.} There has been a line of research, called neural-symbolic computing (NSC), that pursues the integration of the symbolic and statistical
paradigms of cognition~\cite{garcez2019neural,kautz2022third,wang2022towards}. NSC has a long history, dating back to McCulloch and Pitts 1943 paper$_{\!}$~\cite{mcculloch1943logical}, even before AI~was
recognized$_{\!}$ as$_{\!}$ a$_{\!}$ new$_{\!}$ scientific$_{\!}$ field.$_{\!}$ During 2000s, NSC$_{\!}$ received systematic$_{\!}$ study~\cite{pollack1990recursive,shastri1993simple,towell1994knowledge,plate1995holographic}.$_{\!}$ Early NSC systems were meticulously designed for hard logic reasoning, but they~are far$_{\!}$~less$_{\!}$~trainable,$_{\!}$~and$_{\!}$~fall$_{\!}~$short$_{\!}~$when$_{\!}~$solving$_{\!}~$real-world$_{\!}$~pro-\\
\noindent blems. NSC$_{\!}$ has$_{\!}$ recently$_{\!}$ ushered in its renaissance, since it shows promise of reconciling statistical learning of neural networks and logic reasoning of abstract knowledge -- which is viewed as a key enable to the next
generation of AI$_{\!}$~\cite{lake2017building,marcus2018deep}.$_{\!}$ Specifically,$_{\!}$ recent$_{\!}$ NSC$_{\!}$ systems$_{\!}$~\cite{giunchiglia2021multi,giunchiglia2022deep}$_{\!}$ show the possibility for modern neural networks to manipulate abstract knowledge with diverse forms of symbolic representation, including knowledge graph\!~\cite{cranmer2020discovering,teru2020inductive,lin2019kagnet}, propositional logic\!~\cite{hu2016harnessing,stewart2017label,xu2018semantic}, and first-order$_{\!}$ logic\!~\cite{rocktaschel2017end,diligenti2017semantic,fischer2019dl2}.$_{\!}$ They$_{\!}$ also$_{\!}$ demonstrate successful application in several domains and disciplines, \eg, scientific discovery\!~\cite{segler2018planning,dai2019retrosynthesis}, program generation\!~\cite{valkov2018houdini,nye2020learning,parisotto2017neuro}, (visual) question-answering\!~\cite{vedantam2019probabilistic,yi2018neural}, robot planning\!~\cite{mou2017coupling,lyu2019sdrl,yang2018peorl}, and  mathematical reasoning\!~\cite{arabshahi2018combining,li2020closed,lample2019deep}.

To date, none of NSC systems reports advanced performance in large-scale vision, to our best knowledge. In this work, we take the lead to promote and implement the idea of conciliating the methodologies of symbolic and neural paradigms,$_{\!}$ in$_{\!}$ visual$_{\!}$ semantic$_{\!}$ interpretation.$_{\!}$ Moreover,$_{\!}$ many previous NSC systems$_{\!}$ only$_{\!}$ exploit$_{\!}$ logical$_{\!}$ constraints during network training$_{\!}$~\cite{xu2018semantic,manhaeve2018deepproblog,dai2019bridging,xie2019embedding,fischer2019dl2,badreddine2022logic},  while our solution is more favored as logic rules are involved throughout network training$_{\!}$ and$_{\!}$ inference.$_{\!}$ As$_{\!}$ a$_{\!}$ result,$_{\!}$ impressive$_{\!}$ performances across diverse challenging datasets are delivered, and in turn, provide solid empirical evidence for the power of NSC.

\section{Methodology}
\label{sec:method}

\noindent\textbf{Task$_{\!}$ Setup$_{\!}$ and$_{\!}$ Notations.$_{\!}$} In$_{\!}$ this$_{\!}$ work$_{\!}$ we$_{\!}$ are$_{\!}$ interested$_{\!}$~in structured$_{\!}$ visual$_{\!}$ parsing$_{\!}$~\cite{li2022deep}$_{\!}$ --$_{\!}$ a$_{\!}$ more$_{\!}$ challenging$_{\!}$ yet$_{\!}$ realis- tic$_{\!}$ setting$_{\!}$ for$_{\!}$ semantic$_{\!}$ segmentation$_{\!}$ --$_{\!}$ where$_{\!}$ both$_{\!}$ semantic concepts and their relations are considered in a form of a tree-shaped$_{\!}$ class$_{\!}$ hierarchy$_{\!}$ $\mathcal{T}\!=\!\langle\mathcal{V}, \mathcal{E}\rangle$$_{\!}$.$_{\!}$ The$_{\!}$ node$_{\!}$ set$_{\!}$ $\mathcal{V}\!=\!\cup^L_{l=1\!}\mathcal{V}_l$ represents the classes/concepts at $L$ abstraction levels. For instance in Fig.$_{\!}$~\ref{fig:2}(a), the leaf nodes $\mathcal{V}_1$ are the finest classes (\eg, \texttt{chair},  \texttt{pot}), while the internal nodes are higher-level concepts~(\eg, \texttt{furniture},$_{\!}$ \texttt{utensil}),$_{\!}$ and$_{\!}$ the$_{\!}$ roots$_{\!}$ $\mathcal{V}_{L\!}$ are$_{\!}$ the$_{\!}$ most$_{\!}$ abstract$_{\!}$ ones$_{\!}$ (\eg,$_{\!}$ \texttt{object}).$_{\!}$ The$_{\!}$ edge$_{\!}$ set$_{\!}$ $\mathcal{E}_{\!}$ encodes$_{\!}$~relational knowledge$_{\!}$ among$_{\!}$ classes.$_{\!}$ For$_{\!}$ example,$_{\!}$ a directed$_{\!}$ edge$_{\!}$ $u_{\!\!}\rightarrow_{\!\!}v\!\in\!\mathcal{E}$$_{\!}$~denotes a \textit{part-of} relation between classes $u,v\!\in\!\mathcal{V}$ in \textit{adjacent} levels (\eg,$_{\!}$ \texttt{utensil}$\rightarrow$\texttt{pot}). 

Given $\mathcal{T}$, the$_{\!}$ target$_{\!}$ goal$_{\!}$ is$_{\!}$ to$_{\!}$ assign$_{\!}$ each$_{\!}$ pixel
a \textit{valid} root-to-leaf path in $\mathcal{T}$. For instance, associating a pixel with \texttt{object}$\rightarrow$\texttt{utensil}$\rightarrow$\texttt{pot} is valid, yet with \texttt{object}$\rightarrow$\\
\noindent \texttt{furniture}$\rightarrow$\texttt{pot} is \textit{invalid}. Thus standard semantic segmentation can be viewed as a specific case of such structured setting --- only assigning pixels with one single class label from the leaf nodes $\mathcal{V}_1$ without considering the hierarchy.

\begin{figure*}[t]
  \vspace{-2pt}
  \begin{center}
    \includegraphics[width=\linewidth]{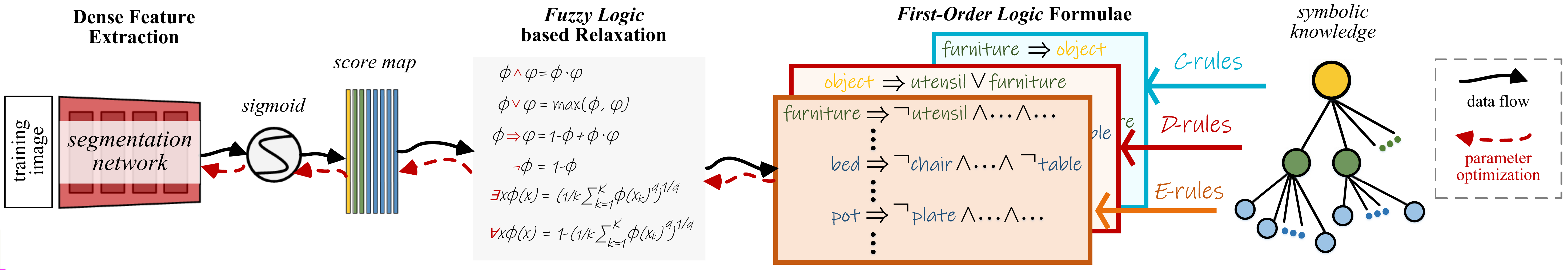}
  \end{center}
  \vspace{-13pt}
  \captionsetup{font=small}
  \caption{\small{Illustration of our logic-induced network training (\S\ref{sec:training}). For clarity, the pixel-wise binary cross-entropy loss is omitted.}}
  \vspace{-12pt}
  \label{fig:3}
\end{figure*}

\noindent\textbf{Algorithmic$_{\!}$ Overview.$_{\!}$} \textsc{LogicSeg}$_{\!}$ is$_{\!}$ a$_{\!}$ unified,$_{\!}$ neural-logic learning and reasoning model for visual parsing, supported by$_{\!}$ large-scale$_{\!}$ data$_{\!}$ and$_{\!}$ the$_{\!}$ structured$_{\!}$ symbolic$_{\!}$ knowledge$_{\!}$ $\mathcal{T}$.
\begin{itemize}[leftmargin=*]
  \setlength{\itemsep}{0pt}
  \setlength{\parsep}{-2pt}
  \setlength{\parskip}{-0pt}
  \setlength{\leftmargin}{-10pt}
  \vspace{-6pt}
   \item From the \textit{\textbf{neural}} aspect, \textsc{LogicSeg} is \textit{model-agnostic}.~After dense feature extraction, its classification head outputs a total of $|\mathcal{V}|$ \textit{sigmoid}-normalized scores, \ie, $\bm{s}\!\in\![0,1]^{|\mathcal{V}|}$, over$_{\!}$ all$_{\!}$ the$_{\!}$ classes$_{\!}$ $\mathcal{V}$~for$_{\!}$ each$_{\!}$ pixel,$_{\!}$ like$_{\!}$~\cite{li2022deep}.$_{\!}$ Here$_{\!}$ $|_{\!}\cdot_{\!}|_{\!}$ counts its$_{\!}$ elements.$_{\!}$ A$_{\!}$ set$_{\!}$ of$_{\!}$ logic$_{\!}$ rules,$_{\!}$ derived$_{\!}$ from$_{\!}$ $\mathcal{T}$, are injected into network training and inference.
   \item From the \textit{\textbf{logic}} aspect, \textsc{LogicSeg} uses \textit{first-order logic} to express the complex and abstract relational knowledge in$_{\!}$ $\mathcal{T}$.$_{\!}$ The$_{\!}$ network$_{\!}$ is$_{\!}$ learnt$_{\!}$ as$_{\!}$ approximation$_{\!}$ of$_{\!}$ logic$_{\!}$~predi-  cates by following the logical specifications. Once trained, it conducts iterative reasoning on the basis of logic rules.
  \vspace{-4pt}
\end{itemize}
After introducing our logic based visual relational knowledge representation (\S\ref{sec:knolwedge}), we will elaborate on our logic-induced network training (\S\ref{sec:training}) and inference (\S\ref{sec:inference}) strategies.

\subsection{Parsing Visual Semantics with Logic Rules}\label{sec:knolwedge}
We formalize our target task --- \textit{learning and reasoning vi- sual$_{\!}$ semantics$_{\!}$ with$_{\!}$ logic}$_{\!}$ ---$_{\!}$ as$_{\!}$ a$_{\!}$ triple$_{\!}$ $\langle\mathcal{T},_{\!} \mathcal{X},_{\!} \Pi\rangle$.$_{\!}$ $\mathcal{X}$$_{\!}$~is$_{\!}$~a$_{\!}$ data collection,$_{\!}$ \ie,$_{\!}$ $\mathcal{X}\!=\!\{(x_k,\bm{y}_k)\}_{k=1}^K$,$_{\!}$ where$_{\!}$ $x_{k\!}$ is$_{\!}$ a$_{\!}$ pixel$_{\!}$ data point,$_{\!}$ and$_{\!}$ $\bm{y}_{k\!}\!\in_{\!}\!\{0,1\}^{|\mathcal{V}|\!}$ is$_{\!}$ its$_{\!}$ groundtruth$_{\!}$ symbolic$_{\!}$~descrip- tion$_{\!}$ in$_{\!}$ terms$_{\!}$ of$_{\!}$ the$_{\!}$ semantic$_{\!}$ hierarchy$_{\!}$ $\mathcal{T}$.$_{\!}$ $\Pi_{\!}$ is$_{\!}$ a$_{\!}$ set$_{\!}$ of$_{\!}$ hierar- chy$_{\!}$ rules$_{\!}$ declaratively$_{\!}$ expressed$_{\!}$ by$_{\!}$ \textit{first-order$_{\!}$ logic},$_{\!}$ contain- ing$_{\!}$  \textbf{i)}$_{\!}$ \textit{constants},$_{\!}$ \eg,$_{\!}$ pixel samples$_{\!}$ $x_1, x_2, \cdots$;$_{\!}$ \textbf{ii)}$_{\!}$ \textit{variables}$_{\!}$ ranging$_{\!}$ over$_{\!}$ constants,$_{\!}$ \eg,$_{\!}$ $x$; and \textbf{iii)} \textit{unary predicates}, one for each class $v\!\in\!\mathcal{V}$, denote the semantics of variables and return \textit{true} and \textit{false}, \eg, $\texttt{bed}(x)\!=\!\textit{true}$ states the fact that pixel$_{\!}$~$x_{\!}$
belongs$_{\!}$ to$_{\!}$ a$_{\!}$ bed.$_{\!}$ A$_{\!}$ logic$_{\!}$ rule/formula$_{\!}$ is$_{\!}$ a$_{\!}$ sequence$_{\!}$~of$_{\!}$ finite$_{\!}$ {predicates}$_{\!}$ with$_{\!}$ \textit{connectives}$_{\!}$ (\ie,$_{\!}$ $\land$,$_{\!}$ $\lor$,$_{\!}$ $\neg$,$_{\!}$ $\Rightarrow$)$_{\!}$ and$_{\!}$ \textit{quan- tifiers} (\ie, $\forall$, $\exists$), and organized in \textit{prenex} form in our case. 

Concretely, $\Pi$ is composed of three types of rules, \ie, \textit{composition}, \textit{decomposition}, and \textit{exclusion}, for comprehensively describing the structured symbolic knowledge $\mathcal{T}$.

\noindent $\bullet$ \textbf{\emph{Composition Rule}} (\textit{C}-rule) expresses our knowledge about the \textit{composition} relations between semantic concepts, such as$_{\!}$ ``\textit{bed$_{\!}$ and$_{\!}$ chair$_{\!}$ are$_{\!}$ (subclasses$_{\!}$ of)$_{\!}$ furniture},''$_{\!}$ in$_{\!}$ a$_{\!}$ form$_{\!}$ of:
\begin{equation}\label{example:C}
\begin{aligned}\small
\small\forall x (\texttt{bed}(x)~&\Rightarrow \texttt{furniture}(x)),\\
\small\forall x (\texttt{chair}(x)&\Rightarrow \texttt{furniture}(x)),
\end{aligned}
\vspace{-2pt}
\end{equation}
where $\texttt{bed}$, $\texttt{chair}$, $\texttt{furniture}$ are predicates, and `$\phi\Rightarrow \varphi$' indicates $\varphi$ is a logical consequence of antecedence $\phi$.
\vspace{-5pt}
\begin{definition}[\textit{C}-rule]
\textit{If one class is labeled true, its superclass should be labeled true} (Fig.$_{\!}$~\ref{fig:2}(b)):
\vspace{-4pt}
\begin{equation}\label{definition:C}
\begin{aligned}\small
\forall x(v(x)\Rightarrow p_v(x)),
  \end{aligned}
  \vspace{-2pt}
\end{equation}
where$_{\!}$ $p_{v\!}$ is$_{\!}$ the$_{\!}$ parent$_{\!}$ node$_{\!}$ of$_{\!}$ $v_{\!}$ in$_{\!}$ $\mathcal{T}$,$_{\!}$ \ie,$_{\!}$ $p_{v\!}\!\rightarrow\!v\!\in\!\mathcal{E}_{\!}$ (the$_{\!}$ tree structure$_{\!}$ of$_{\!}$  $\mathcal{T}_{\!}$ restricts$_{\!}$ each$_{\!}$ class$_{\!}$ to$_{\!}$ possess$_{\!}$ only$_{\!}$ one$_{\!}$ super- class).$_{\!}$ \textit{C}-rule$_{\!}$ generalizes$_{\!}$ the$_{\!}$ famous$_{\!}$ \textit{tree-property}\!~\cite{bi2011multilabel,giunchiglia2020coherent}.
  \vspace{-12pt}
\end{definition}
\noindent$\bullet$$_{\!}$ \textbf{\emph{Decomposition$_{\!}$ Rule$_{\!}$}} (\textit{D}-rule)$_{\!}$ states$_{\!}$ our$_{\!}$ knowledge$_{\!}$ about the$_{\!}$ \textit{decomposition}$_{\!}$ relations among$_{\!}$ semantic$_{\!}$ concepts,$_{\!}$ such as$_{\!}$ ``\textit{furniture$_{\!}$ is$_{\!}$ the$_{\!}$ superclass$_{\!}$ of$_{\!}$ bed,$_{\!}$ chair,$_{\!}$ $\cdots$,$_{\!}$ table},''$_{\!}$ via:
\begin{equation}\label{example:D}
\begin{aligned}\small
\forall x (\texttt{furniture}(x)\Rightarrow&\texttt{bed}(x)\lor\texttt{chair}(x)_{\!}~\lor\\
&\cdots\lor\texttt{tabel}(x)).
\end{aligned}
\vspace{-2pt}
\end{equation}
\begin{definition}[\textit{D}-rule]
\textit{$_{\!\!}$If$_{\!}$ one$_{\!}$ class$_{\!}$ is$_{\!}$ labeled$_{\!}$ true,$_{\!}$ at$_{\!}$ least$_{\!}$ one of its {subclasses} should be labeled true}  (Fig.$_{\!}$~\ref{fig:2}(c)):
\vspace{-4pt}
\begin{equation}\label{definition:D}
\begin{aligned}\small
\forall x(v(x)\Rightarrow c^1_v(x)\lor c^2_v(x)\lor\cdots\lor c^N_v(x)),
  \end{aligned}
  \vspace{-2pt}
\end{equation}
where$_{\!}$ $c^n_v\!\in\!\mathcal{C}_{v{\!}}$ are$_{\!}$ all$_{\!}$ the$_{\!}$ child$_{\!}$ nodes$_{\!}$ of$_{\!}$ $v_{\!}$ in$_{\!}$ $\mathcal{T}$,$_{\!}$ \ie,$_{\!}$ $v\!\rightarrow\!c^n_v\!\in\!\mathcal{E}$. \textit{C}-rule and \textit{D}-rule are not equivalent. For instance in Eq.~\ref{example:C}, $\texttt{bed}(x)$ is sufficient but not necessary~for $\texttt{furniture}(x)$: given$_{\!}$ the$_{\!}$ fact$_{\!}$ ``$x_{\!}$ is$_{\!}$ furniture'',$_{\!}$ we$_{\!}$ cannot$_{\!}$ conclude$_{\!}$ ``$x_{\!}$ is$_{\!}$ bed''.
  \vspace{-2pt}
\end{definition}

\noindent $\bullet$ \textbf{\textit{Exclusion Rule}} (\textit{E}-rule) specifies our knowledge about \textit{mutual exclusion} relations between \textit{sibling} concepts, such as ``\textit{a bed cannot be at the same time a chair},'' in a form of:
\vspace{-2pt}
\begin{equation}\label{example:E}
\begin{aligned}\small
\forall x (\texttt{bed}(x)\Rightarrow \neg \texttt{chair}(x)).
\end{aligned}
\vspace{-2pt}
\end{equation}
\begin{definition}[\textit{E}-rule]
\textit{If one class is labeled true, all its sibling {classes} should be labeled false}  (Fig.$_{\!}$~\ref{fig:2}(d)):
\vspace{-4pt}
\begin{equation}\label{definition:E}
\begin{aligned}\small
\forall x(v(x)\Rightarrow \neg a^1_v(x)\land \neg a^2_v(x)\land\cdots\land \neg a^{M}_v(x)),
  \end{aligned}
  \vspace{-2pt}
\end{equation}
where$_{\!}$ $a^{m\!}_{v\!}\!\in\!\mathcal{A}_{v\!}$ are$_{\!}$ all$_{\!}$ the$_{\!}$ peer$_{\!}$ nodes$_{\!}$ of$_{\!}$ $v_{\!}$ in$_{\!}$ $\mathcal{T}$$_{\!}$.$_{\!}$ Note$_{\!}$ that$_{\!}$ \textit{E}-rule is$_{\!}$~ignored$_{\!}$ by$_{\!}$ many$_{\!}$ hierarchy-aware$_{\!}$ algorithms\!~\cite{li2022deep,bertinetto2020making,wehrmann2018hierarchical}.
\end{definition}

\subsection{Logic-Induced Training}\label{sec:training}
So$_{\!}$ far,$_{\!}$ we$_{\!}$ shown$_{\!}$ the$_{\!}$ logic$_{\!}$ rules$_{\!}$ $\Pi$$_{\!}$ provide$_{\!}$ \textsc{LogicSeg} a$_{\!}$~flexible$_{\!}$ language$_{\!}$ for$_{\!}$ comprehensively$_{\!}$ expressing$_{\!}$ the$_{\!}$ complex \textit{meronymy}
and \textit{exclusion} relations among symbolic concepts in the hierarchy $\mathcal{T}$. Unfortunately, these$_{\!}$ rules$_{\!}$ are$_{\!}$ logic$_{\!}$ formulae$_{\!}$ working$_{\!}$ with$_{\!}$ variables$_{\!}$ (assuming a boolean value), and non-differentiable logic symbols (\eg, $\forall$, $\Rightarrow$). This prevents the integration with end-to-end network learning.

\begin{figure*}[t]
   \vspace{-10pt}
   \begin{center}
      \includegraphics[width=\linewidth]{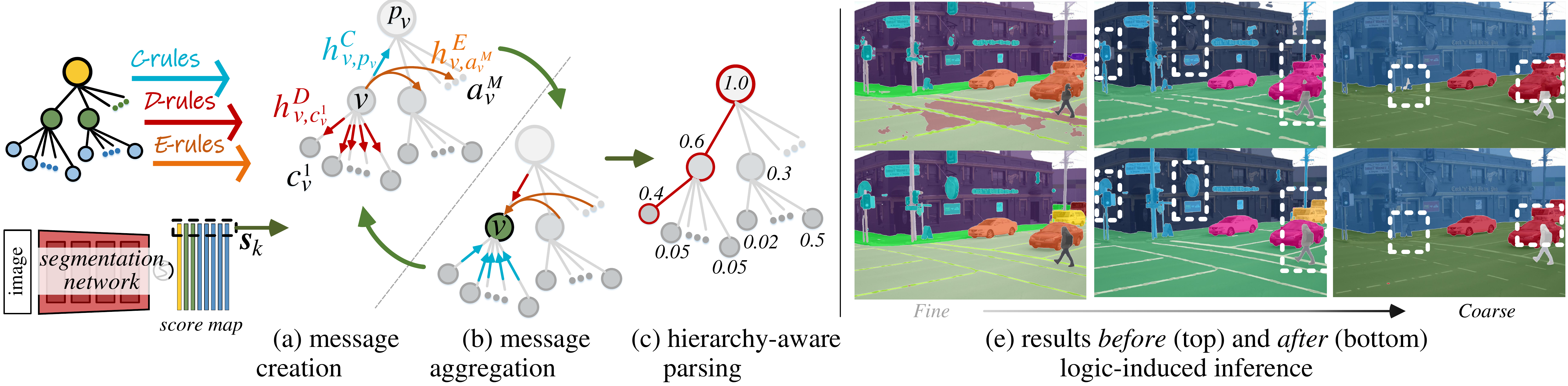}
    \put(-388.2,3.02){\scriptsize (Eq.\!~\ref{eq:cm})}
    \put(-319.8,3.02){\scriptsize (Eq.\!~\ref{eq:iu})}
    \put(-260.8,3.02){\scriptsize (Eq.\!~\ref{eq:infer})}
   \end{center}
   \vspace{-15pt}
   \captionsetup{font=small}
   \caption{\small{Illustration of our logic-induced inference (\S\ref{sec:inference}). (a-b) Iterative reasoning is made by exchanging and absorbing messages between nodes, following the logic rules $\Pi$.  For clarity, we only show the message creation (Eq.~\ref{eq:cm}) and aggregation (Eq.~\ref{eq:iu}) stages for one single node. (c) Structured parsing (Eq.~\ref{eq:infer}) is conducted by selecting the top-scoring path $\mathcal{P}^{*\!}$ (highlighted in red) after logic-guided iterative reasoning. (d) With logic-induced inference, \textsc{LogicSeg} is able to generate more accurate and hierarchy-compliant predictions.}}
   \vspace{-12pt}
   \label{fig:4}
\end{figure*}

Inspired by~\cite{van2022analyzing,badreddine2022logic}, a \textit{fuzzy logic} based \textit{grounding} process$_{\!}$ is$_{\!}$ adopted$_{\!}$ to$_{\!}$ interpret$_{\!}$  logic$_{\!}$ formulae$_{\!}$ as$_{\!}$ differentiable fuzzy$_{\!}$ relations$_{\!}$ on$_{\!}$ real$_{\!}$ numbers$_{\!}$ for$_{\!}$ neural$_{\!}$ computing$_{\!}$ (Fig.$_{\!}$~\ref{fig:3}).

\noindent\textbf{Fuzzy {relaxation}.} Fuzzy logic is a form of soft probabilistic logic. It deals with reasoning that is approximate instead of fixed and exact; variables have a truth degree that ranges in $[0, 1]$: zero and one meaning that the variable is \textit{false} and \textit{true} with certainty, respectively\!~\cite{novak2012mathematical}. Hence we can ground predicates onto segmentation network outputs. For instance, given a pixel sample $x$, corresponding network prediction score \wrt class \textit{bed} is a grounded predicate \wrt $\texttt{bed}(x)$.
{Logical connectives}, \ie, $\land, \lor, \neg, \Rightarrow$ are approximated with \textit{fuzzy operators}, \ie, \textit{t-norm}, \textit{t-conorm}, \textit{fuzzy negation}, and \textit{fuzzy implication}. As suggested by$_{\!}$~\cite{van2022analyzing}, we adopt the operators in \textit{Goguen fuzzy logic}\!~\cite{hajek2013metamathematics} and \textit{Gödel fuzzy logic}\!~\cite{feferman1986kurt}:
\vspace{-2pt}
\begin{equation}\label{eq:fc}
\begin{aligned}\small
\phi\land \varphi&= \phi\cdot \varphi, \ \ \  \ ~~\phi \lor \varphi= \texttt{max}(\phi, \varphi),\\
\neg \phi &= 1-\phi, \ \ \ \ \phi\Rightarrow \varphi= 1-\phi+\phi\cdot \varphi.
  \end{aligned}
    \vspace{2pt}
\end{equation}
The existential quantifier $\exists$ and universal quantifier $\forall$ are

\noindent approximated in a form of generalized mean:
\vspace{-4pt}
\begin{equation}\label{eq:fq}
\begin{aligned}\small
\exists x\phi(x)=~& \textstyle(\frac{1}{K}\!\textstyle\sum\nolimits_{k=1}^K \phi(x_k)^q)^\frac{1}{q}, \\
\forall x\phi(x)=~& 1\!-\!\textstyle(\frac{1}{K}\!\textstyle\sum\nolimits_{k=1}^K (1\!-\!\phi(x_k))^q)^\frac{1}{q},
  \end{aligned}
\end{equation}
where $q\!\in\!\mathbb{Z}$. Please refer to$_{\!}$~\cite{van2022analyzing,badreddine2022logic} for detailed discussion regarding$_{\!}$ the$_{\!}$ rationale$_{\!}$ behind$_{\!}$ such$_{\!}$ approximation$_{\!}$ of$_{\!}$ $\exists_{\!}$ and$_{\!}$ $\forall$.

\noindent\textbf{Logic Loss.} With fuzzy {relaxation}, we are ready to convert our first-order logic rules $\Pi$ into loss functions. 

\noindent $\bullet$ \textbf{\emph{\textit{C}-rule Loss.}} For a non-root node $v\!\in\!\mathcal{V}/\mathcal{V}_L$, its corres- ponding \textit{C}-rule$_{\!}$ (\textit{cf}.$_{\!}$~Eq.$_{\!\!}$~\ref{definition:C}) is grounded as:
\vspace{-5pt}
\begin{equation}\label{eq:CL1}
\begin{aligned}\small
\mathcal{G}_C(v)\!=1\!-\!\textstyle\Big(\frac{1}{K}\!\sum_{k=1}^K(\bm{s}_k[v] \!-\! \bm{s}_k[v]\cdot\bm{s}_k[p_{v}])^{q}\Big)^{\!\frac{1}{q}},
  \end{aligned}
    \vspace{-3pt}
\end{equation}
where $\bm{s}_k[v]\!\in\![0,1]$ refers to the score (confidence) of $x_k$ for class $v$. Then the \textit{C}-rule based training objective is given as:
\vspace{-5pt}
\begin{equation}\label{eq:CL2}
\begin{aligned}\small
\mathcal{L}_C\!=\!\textstyle\frac{1}{|\mathcal{V}|-|\mathcal{V}_L|}\!\!\textstyle\sum_{v\in\mathcal{V}/\mathcal{V}_L}\!1-\mathcal{G}_C(v).
  \end{aligned}
      \vspace{2pt}
\end{equation}

\noindent $\bullet$ \textbf{\emph{\textit{D}-rule Loss.}} For a non-leaf node $v\!\in\!\mathcal{V}/\mathcal{V}_1$, its corres- ponding \textit{D}-rule$_{\!}$ (\textit{cf}.$_{\!}$~Eq.$_{\!\!}$~\ref{definition:D}) is grounded as:
\vspace{-5pt}
\begin{equation}\label{eq:DL1}
\begin{aligned}\small
\!\!\!\!\!\!\!\!\mathcal{G}_D(v)\!=\!1\!-\!\textstyle\Big(\!\frac{1}{K}\!\!\textstyle\sum\limits_{k=1}^K\!\!\big(\bm{s}_k[v] \!-\! \bm{s}_k[v]\!\cdot\!\texttt{max}(\{\bm{s}_k[c^n_v]\}_n)\big)^{\!q}\Big)^{\!\frac{1}{q}}.\!\!\!\!
  \end{aligned}
    \vspace{-5pt}
\end{equation}
Similarly, our \textit{D}-rule loss is given as:
\vspace{-3pt}
\begin{equation}\label{eq:DL2}
\begin{aligned}\small
\mathcal{L}_D\!=\!\textstyle\frac{1}{|\mathcal{V}|-|\mathcal{V}_1|}\!\!\textstyle\sum\nolimits_{v\in\mathcal{V}/\mathcal{V}_1}\!1-\mathcal{G}_D(v).
  \end{aligned}
    \vspace{-2pt}
\end{equation}

\noindent $\bullet$ \textbf{\emph{\textit{E}-rule Loss.}} During the grounding of \textit{E}-rule$_{\!}$ (\textit{cf}.$_{\!}$~Eq.$_{\!\!}$~\ref{definition:E}), we first translate such \textit{one}-\textit{vs}-\textit{all} exclusion statement to a semantically equivalent expression, \ie, the aggregation of multiple \textit{one}-\textit{vs}-\textit{one} exclusion ($\{(v(x)_{\!}\Rightarrow_{\!}\neg a^{1}_v(x)), \cdots\!, \{(v(x)_{\!}\Rightarrow_{\!}\neg a^{M}_v(x))\}$). Adopting such translation is to avoid the \textit{sorites paradox}, \ie, {a long chain of only slightly unreliable deductions can be very unreliable}\!~\cite{goguen1969logic} (\eg, $0.9^{10}\!\approx\!0.34$), happened during approximating a series of $\land$. Then, for each node $v\!\in\!\mathcal{V}$, its corresponding \textit{E}-rule$_{\!}$ is grounded as:
\vspace{-6pt}
\begin{equation}\label{eq:EL1}
\begin{aligned}\small
\!\!\!\!\!\!\!\!\mathcal{G}_E(v)\!=\!1\!-\!\textstyle\frac{1}{M}\!\!\textstyle\sum_{m=1}^M\!\!\Big(\!\frac{1}{K}\!\!\textstyle\sum_{k=1}^K\!\!\big(\bm{s}_k[v]\!\cdot\!\bm{s}_k[a^m_v]\big)^{\!q}\Big)^{\!\frac{1}{q}}.\!\!\!\!
  \end{aligned}
    \vspace{-3pt}
\end{equation}
Similarly, our \textit{E}-rule loss is given as:
\vspace{-4pt}
\begin{equation}\label{eq:EL2}
\begin{aligned}\small
\mathcal{L}_E\!=\!\textstyle\frac{1}{|\mathcal{V}|}\!\!\sum\nolimits_{v\in\mathcal{V}}1-\mathcal{G}_E(v).
  \end{aligned}
    \vspace{-3pt}
\end{equation}
In this way, it is possible to backpropagate the gradient from

\noindent the logic loss into the
network. The network is essentially learned as neural predicates obeying the logical constraints. It is worth mentioning that, due to large-scale training, it is infeasible to compute the full semantics of $\forall$; batch-training can be viewed as sampling based approximation~\cite{van2022analyzing}.

Our overall training target is organized as:
    \vspace{-4pt}
\begin{equation}\label{eq:AL}
\begin{aligned}\small
\mathcal{L}\!=\!\alpha(\mathcal{L}_C\!+\!\mathcal{L}_D\!+\!\mathcal{L}_E)\!+\!\textstyle\frac{1}{K}\!\!\textstyle\sum\nolimits_{k=1\!}^K\mathcal{L}_{\text{BCE}}(\bm{s}_k, \bm{y}_k).
  \end{aligned}
    \vspace{-4pt}
\end{equation}
Here$_{\!}$ $\bm{y}\!\in\!\{0,_{\!}1\}^{|\mathcal{V}|\!}$ is$_{\!}$ the$_{\!}$ groundtruth,$_{\!}$ $\mathcal{L}_{\text{BCE}\!}$ is$_{\!}$ the$_{\!}$ binary$_{\!}$ cross- entropy$_{\!}$ loss,$_{\!}$ and$_{\!}$ the$_{\!}$ coefficient$_{\!}$ is$_{\!}$ empirically$_{\!}$ set$_{\!}$ as$_{\!}$ $\alpha\!=\!0.2$.

\subsection{Logic-Induced Inference}\label{sec:inference}

We just showed that \textsc{LogicSeg} can approximate the~pre- dicates by integrating symbolic logic constraints  into large-scale network training. However, during inference, there is {no} explicit way to ensure the alignment between the class hierarchy $\mathcal{T}$ and network prediction, {neither} sound reasoning with the logic rules $\Pi$. We thus put forward \textit{logic-induced$_{\!}$ reasoning}$_{\!}$ (Fig.$_{\!}$~\ref{fig:4}),$_{\!}$ where$_{\!}$ the$_{\!}$ logic$_{\!}$ rules$_{\!}$ $\Pi_{\!}$ are$_{\!}$~encapsulated into an iterative optimization process. Such process is non-learnable, based on only matrix operations and thus can be seamlessly embedded into network feed-forward inference, yielding an elegant yet compact neural-logic visual parser.

Our solution is built upon the classic \textit{message passing} algorithm which is to estimate the marginal likelihood for a given tree structure by \textit{iteratively} exchanging messages between nodes.$_{\!}$ Specifically,$_{\!}$ at$_{\!}$ each$_{\!}$ iteration,$_{\!}$ for$_{\!}$ each$_{\!}$ pixel sample $x_k$, node $v\!\in\!\mathcal{V}$ sends different types of messages to different neighboring nodes, according to the logic rules $\Pi$:
\vspace{-5pt}
\begin{equation}
\begin{aligned}
\small\!\!\!\!\!\!\text{\textit{C}-message}\!:\!~h^C_{v, p_v\!}&\!=\!v(x_k)\Rightarrow p_v(x_k)\\
&\!=\!1\!-\!\bm{s}_k[v]\!+\! \bm{s}_k[v]\!\cdot_{\!}\bm{s}_k[p_{v}],~~~~~~~~~~~~~~~~~~~~~\nonumber
\end{aligned}
 \vspace{-4pt}
\end{equation}
\begin{equation}
\begin{aligned}\label{eq:cm}
\small\!\!\!\!\!\!\text{\textit{D}-message}\!:\!~h^D_{v, c^n_v\!}&\!=\!v(x_k)\!\Rightarrow\! c^1_v(x_k)\!\lor\!\cdots\!\lor\! c^N_v(x_k)\\
&\!=\!1\!-\!\bm{s}_k[v]\!+\! \bm{s}_k[v]\!\cdot\!\texttt{max}(\{\bm{s}_k[c^n_v]\}_n),
\end{aligned}
 \vspace{-2pt}
\end{equation}
\begin{equation}
\begin{aligned}
\small\!\!\!\!\!\!\!\!\!\!\text{\textit{E}-message}\!:_{\!}\!~h^E_{v, a^m_v\!}&\!=\!-1{\!}\cdot{\!}\big(v(x_k)\!\Rightarrow{\!} \neg a^1_v(x_k)\!\land\!\cdots\!\land\! \neg a^{M\!}_v(x_k)\big)\!\!\!\!\!\!\!\!\\
&\!=\!-_{\!}~(1\!-\!\textstyle\frac{1}{M}_{\!}\!\sum_{m=1}^M\!\bm{s}_k[v]_{\!}\cdot_{\!}\bm{s}_k[a_{v}^m]).\nonumber\!\!\!\!
\end{aligned}
\end{equation}
Node $v$ is updated by aggregating the received messages:
\vspace{-3pt}
\begin{equation}\label{eq:iu}
\begin{aligned}\small
\!\!\!\!\!\!\bm{s}_k[v]\!\leftarrow\!\bm{s}_k[v]&\!+\!\textstyle\frac{1}{N}\!\!\sum\nolimits_{c^n_v\in\mathcal{C}_v\!\!}\bm{s}_{k}[c^n_v]\!\cdot\! h^C_{c^n_v, v}\!+\!\bm{s}_{k}[p_v]\!\cdot\! h^D_{p_v, v}\!\!\!\\
&\!+\!\textstyle\frac{1}{M}\!\!\sum\nolimits_{a^m_v\in\mathcal{A}_v\!\!}\bm{s}_{k}[a^m_v]\!\cdot\! h^E_{a^m_v, v}.
  \end{aligned}
\end{equation}
Each message (\textit{cf}.$_{\!}$~Eq.$_{\!}$~\ref{eq:cm}) accounts for the certainty degree that $v$ satisfies the corresponding logic rule (\textit{cf}.$_{\!}$~\S\ref{sec:knolwedge}) when being$_{\!}$  grounded$_{\!}$  on$_{\!}$  pixel$_{\!}$  data$_{\!}$  point$_{\!}$  $x_k$,$_{\!}$  with$_{\!}$  fuzzy$_{\!}$ logic$_{\!}$ based approximation$_{\!}$ (\textit{cf}.\!~\S\ref{sec:training}).$_{\!}$ Intuitively,$_{\!}$ the$_{\!}$ more$_{\!}$ certainty$_{\!}$ a$_{\!}$ node meets the logic rules, the more message it can propagate to other nodes. Note that, $v$ creates a \textit{negative} message $h^E_{v, a^m_v\!}$~to  ``suppress'' other peer nodes due to their exclusion relations. In Eq.$_{\!}$~\ref{eq:iu}, the received messages are weighted by the confi- dence of the source nodes themselves -- the grounded predi- cates,$_{\!}$ \ie,$_{\!}$ $\bm{s}_{k}[c^n_v]$,$_{\!}$ $\bm{s}_{k}[p_v]$,$_{\!}$ and$_{\!}$ $\bm{s}_{k}[a^m_v]$.$_{\!}$  After$_{\!}$ each$_{\!}$ iteration, the score vector $\bm{s}_k$ is \textit{softmax}-normalized per hierarchy level. 

Finally, each pixel $x_k$ is associated with the top-scoring \textit{root-to-leaf} path in the hierarchy $\mathcal{T}$  (red path in Fig.~\ref{fig:4}(c)):
\vspace{-4pt}
\begin{equation}
   \begin{aligned}\label{eq:infer}\small
      \mathcal{P}^*=\{v^*_1, \cdots, v^*_{L}\}=\!\argmax_{\mathcal{P}\subset\mathcal{T}}\textstyle\sum\nolimits_{v^\mathcal{P}\in\mathcal{P}}\bm{s}_k[v^\mathcal{P}],
   \end{aligned}
\vspace{-6pt}
\end{equation}
where $\mathcal{P}_{\!}\!=\!\{v^\mathcal{P}_1, \cdots_{\!}, v^\mathcal{P}_{L}\}\!\subset\!\mathcal{T}$~indicates a feasible root-to-leaf path in~$\mathcal{T}$, \ie, $\forall v^{\mathcal{P}\!}_{l}, v^{\mathcal{P}}_{l-1\!}\in\!\mathcal{\mathcal{P}}\!\Rightarrow\!v^{\mathcal{P}\!\!}_{l\!}\!\rightarrow\!v^{P}_{l-1\!}\in\!\mathcal{E}$.

It is easy to find that all the logic-induced inference steps (\textit{cf}.$_{\!}$~Eq.$_{\!}$~\ref{eq:cm}-\ref{eq:infer}) can be formulated in \textit{matrix} form with only a couple of matrix multiplications (see corresponding pseudo-code in the supplementary). Hence it is efficient on GPU and can be straightforward injected into the network, making \textsc{LogicSeg} a fully-integrated neural-logic machine. In practice, $2$-iteration message passing is enough for robust prediction. Through logic-induced reasoning (\textit{cf}.$_{\!}$~Eq.$_{\!}$~\ref{eq:iu}) and hierarchy-aware parsing (\textit{cf}.$_{\!}$~Eq.$_{\!}$~\ref{eq:infer}), \textsc{LogicSeg} is able to \textbf{i)} obtain \textit{improved performance}, and \textbf{ii)} guarantee the parsing results to
\textit{respect the hierarchy} $\mathcal{T}$, with \textbf{iii)} only  \textit{negligible speed delay} (about 3.8\%). See \S\ref{sec:ablation} for related experiments.

\section{Experiment} \label{sec:4}

\subsection{Experimental Setup}
\noindent\textbf{Datasets.}$_{\!}$ We$_{\!}$ conduct$_{\!}$ extensive$_{\!}$ experiments$_{\!}$ on$_{\!}$ four$_{\!}$ datasets, \ie, Mapillary$_{\!}$ Vistas$_{\!}$ 2.0\!~\cite{neuhold2017mapillary}, Cityscapes\!~\cite{cordts2016cityscapes}, Pascal-Part-108\!~\cite{chen2014detect}, and {ADE20K}$_{\!}$~\cite{zhou2017scene}. The four datasets are selected to cover the rich application scenarios of semantic segmentation, including urban street segmentation for automated driving (\ie, \cite{neuhold2017mapillary,cordts2016cityscapes}), object part parsing (\ie, \cite{chen2014detect}), and fine-grained understanding of daily scenes  (\ie, \cite{zhou2017scene}), so as to comprehensively examine the utility of our algorithm.

\begin{itemize}[leftmargin=*]
  \setlength{\itemsep}{0pt}
  \setlength{\parsep}{-2pt}
  \setlength{\parskip}{-0pt}
  \setlength{\leftmargin}{-10pt}
  \vspace{-6pt}
   \item \textbf{Mapillary$_{\!}$ Vistas$_{\!}$ 2.0} is a large-scale urban scene dataset. It contains $18,000$/$2,000/5,000$ images for \texttt{train}/\texttt{val}/ \texttt{test}. A three-level semantic hierarchy, covering $4/16/$ $124$ concepts, is officially provided for dense annotation.
   \item \textbf{Cityscapes} has  $2,975$/$500$/$1,524$ finely annotated, urban street images for \texttt{train}/\texttt{val}/\texttt{test}. The label hierarchy consists of $19$ fine-gained concepts and $6$ superclasses.
   \item \textbf{Pascal-Part-108} is the largest object part parsing dataset. It consists of $4,998$/$5,105$ images for \texttt{train}/\texttt{test}. To establish the class hierarchy, we group $108$ part-level labels into $20$ object-level categories, as in~\cite{michieli2020gmnet,gonzalez2018semantic,zhao2019multi,singh2022float}.
   \item \textbf{ADE20K} is a large-scale generic scene parsing dataset. It is divided into $20,210$/$2,000$/$3,000$ images for \texttt{train}/ \texttt{val}/\texttt{test}. It provides pixel-wise annotations for $150$ fine-grained semantic classes, from which a three-layer label hierarchy (with $3$/$14$/$150$ concepts) can be derived.
  \vspace{-4pt}
\end{itemize}

\noindent\textbf{Evaluation Metric.} We adopt the standard metric, mean intersection-over-union (mIoU), for evaluation. For detailed performance analysis, the score is reported for each hierarchy level $l$ (denoted as mIoU$^l$), as suggested by~\cite{wang2019learning,li2022deep}.

\noindent\textbf{Base Models and Competitors.} To demonstrate our wide benefit, we approach our algorithm on two famous segmentation architectures, \ie, DeepLabV3+~\cite{chen2018encoder} and Mask2For- mer~\cite{cheng2022masked}, with ResNet-101~\cite{he2016deep} and Swin-T~\cite{liu2021swin} backbones. For performance comparison, we involve several hierarchy-aware segmentation models\!~\cite{li2022deep,michieli2020gmnet,singh2022float}, and view \textsc{Hssn}\!~\cite{li2022deep} as our major rival as it is a general framework that reports strong results over several datasets, instead of the others that are dedicated to specific dataset(s) or task setup(s). For comprehensive evaluations, we include a group of previous hierarchy-agnostic segmentation algorithms\!~\cite{yuan2020object,badrinarayanan2017segnet,xiao2018unified,zheng2021rethinking,strudel2021segmenter,xie2021segformer}, whose segmentation results on coarse-grained semantics are obtained by merging the predictions of the corresponding subclasses.

\noindent\textbf{Training.} For the sake of fairness, we follow the standard training setup in~\cite{chen2017deeplab,yuan2020object,cheng2021maskformer,guo2022segnext,lianggmmseg}. In particular, we train $240$K/$80$K iterations for Mapillary Vistas 2.0/Cityscapes, with batch size $8$ and crop size $512$$\times$$1024$; $60$K/$160$K iterations for Pascal-Part-108/ADE20K, with batch size $16$ and crop size $512$$\times$$512$. For data augmentation, the images are horizontally flipped and scaled with a ratio between 0.5$_{\!}$ and$_{\!}$ 2.0$_{\!}$ at$_{\!}$ random.$_{\!}$ For$_{\!}$ network$_{\!}$ optimization,$_{\!}$ SGD (with initial learning rate 1e-2, momentum 0.9, and weight decay 1e-4) and Adam (with initial learning rate 6e-5 and weight decay 0.01) are respectively used for CNN-based and neural attention-based models, where the learning rate is scheduled by the polynomial annealing rule. For network initialization, ImageNet\!~\cite{deng2014large} pre-trained weights are pre-loaded. 

\noindent\textbf{Testing.}$_{\!}$ For$_{\!}$ Mapillary$_{\!}$ Vistas$_{\!}$ 2.0$_{\!}$ and$_{\!}$ Cityscapes,$_{\!}$ we$_{\!}$ keep$_{\!}$ the original$_{\!}$ image$_{\!}$ aspect$_{\!}$ ratio$_{\!}$ but$_{\!}$ resize$_{\!}$ the$_{\!}$ short$_{\!}$ edge$_{\!}$ to$_{\!}$ 1024. Sliding window inference with the identical window$_{\!}$ shape$_{\!}$ as$_{\!}$ the$_{\!}$ training$_{\!}$ size$_{\!}$ is$_{\!}$ adopted$_{\!}$ to$_{\!}$ save$_{\!}$ memory.$_{\!}$ For$_{\!}$ ADE20K$_{\!}$ and

\noindent  Pascal-Part-108,$_{\!}$ the$_{\!}$ short$_{\!}$ edge$_{\!}$ is$_{\!}$ resized$_{\!}$ to$_{\!}$ 512$_{\!}$ so$_{\!}$ as$_{\!}$ to$_{\!}$ enable one-time$_{\!}$ inference$_{\!}$ for$_{\!}$ the$_{\!}$ whole$_{\!}$ image.$_{\!}$ As$_{\!}$ in$_{\!}$ \cite{huang2019ccnet,cheng2021maskformer,yuan2020object,ji2020learning}, performance of all the models is reported  at multiple scales ($\{0.5, 0.75, 1.0, 1.25,$ $ 1.5, 1.75\}_{\!}$) with horizontal flipping.

\noindent\textbf{Hyperparameters.} We set $\alpha\!=\!0.2$ for the loss coefficient (\textit{cf}.\!~Eq.\!~\ref{eq:AL}), and $q\!=\!5$ for logic quantifier approximation (\textit{cf}.\!~Eq.\!~\ref{eq:fq}), as suggested by \cite{badreddine2022logic}. For network inference, we find 2 iterations of message passing are enough.

\begin{table}
  \centering
  \small
  \resizebox{0.99\columnwidth}{!}{
    \setlength\tabcolsep{3.7pt}
    \renewcommand\arraystretch{1.10}
    \begin{tabular}{r|c|ccc}
      \toprule
      {Method} & {Backbone} & $\text{mIoU}^3$$\uparrow$& $\text{mIoU}^2$$\uparrow$& $\text{mIoU}^1$$\uparrow$ \\
      \midrule
      Seamless~\cite{porzi2019seamless} & ResNet-101 & 84.23 & 70.24 &  38.82  \\
      OCRNet~\cite{yuan2020object}
      &  HRNet-W48 & 84.19& 69.82& 38.26  \\
      HMSANet~\cite{wang2020hierarchical}
      & HRNet-W48 & 84.63& 70.71 & 39.53 \\
      Mask2Former~\cite{cheng2022masked}& Swin-S  & 88.81 & 74.98 & 43.49 \\
      \textsc{Hssn}~\cite{li2022deep}
      &ResNet-101  & 85.27 & 71.40 &  40.16\\
      \textsc{Hssn}~\cite{li2022deep}
      &Swin-S   & 90.02 & 75.81 &43.97 \\
      \hline
      DeepLabV3+~\cite{chen2018encoder}
      &  & 81.86 & 68.17 & 37.43 \\
      {+ \textbf{\textsc{LogicSeg}}}
      & \multirow{-2}{*}{ResNet-101}  & \reshl{\textbf{85.51}}{3.65} & \reshl{\textbf{71.69}}{3.42} & \reshl{\textbf{40.72}}{3.29} \\
      \hline
      MaskFormer~\cite{cheng2021maskformer} & & 87.93 & 73.88 & 42.16 \\
      {+ \textbf{\textsc{LogicSeg}}}
      & \multirow{-2}{*}{Swin-S}  &  \reshl{\textbf{90.35}}{2.42}  &  \reshl{\textbf{76.61}}{2.73}  &  \reshl{\textbf{45.12}}{2.96}  \\
      \bottomrule[1.0pt]
    \end{tabular}
  }
  \vspace{-7pt}
  \captionsetup{font=small}
  \caption{\small\textbf{$_{\!}$Urban$_{\!}$ scene$_{\!}$ parsing$_{\!}$ results$_{\!}$} (\S\ref{sec:4.1})$_{\!}$ on$_{\!}$ Mapillary$_{\!\!}$ Vistas$_{\!}$~$2.0_{\!}$ \cite{neuhold2017mapillary} \texttt{val} with a three-level label hierarchy of $4$/$16$/$124$ concepts.}
  \label{table:mapillary}
  \vspace{-7pt}
\end{table}

\begin{table}
  \centering
  \small
  \resizebox{0.87\columnwidth}{!}{
    \setlength\tabcolsep{4.2pt}
    \renewcommand\arraystretch{1.1}
    \begin{tabular}{r|c|cc}
      \toprule[1.0pt]
      {Method} & {Backbone} & $\text{mIoU}^2$$\uparrow$& $\text{mIoU}^1$$\uparrow$ \\
      \midrule
      PSPNet~\cite{zhao2017pyramid} & ResNet-101 & 91.67 & 80.91 \\
      DANet~\cite{fu2019dual}& ResNet-101 & 91.83 &  81.52 \\
      CCNet~\cite{huang2019ccnet} & ResNet-101 & 91.70 &  81.08 \\
      OCRNet~\cite{yuan2020object}&  HRNet-W48 & 92.57& 82.33 \\
      SETR~\cite{zheng2021rethinking} & ViT-L & 92.86 & 82.75  \\
      SegMentor~\cite{strudel2021segmenter} & ViT-L & 91.79 & 81.30  \\
      UperNet~\cite{xiao2018unified}&  Swin-S & 91.92& 81.79 \\
       Mask2Former~\cite{cheng2022masked}  & Swin-S & 93.68 & 83.62\\
      SegFormer~\cite{xie2021segformer} & MiT-B4 &93.81 & 83.90   \\
       \textsc{Hssn}~\cite{li2022deep}  &ResNet-101  & 93.31 & 83.02 \\
      \textsc{Hssn}~\cite{li2022deep} &Swin-S   & 94.39 & 83.74  \\
      \hline
      DeepLabV3+~\cite{chen2018encoder}&   & 92.16 & 82.08  \\
      {+ \textbf{\textsc{LogicSeg}}} & \multirow{-2}{*}{ResNet-101}  & \reshl{\textbf{93.37}}{1.21} & \reshl{\textbf{83.20}}{1.12} \\
      \hline
     MaskFormer~\cite{cheng2021maskformer} &  & 92.96 & 82.57  \\
      {+ \textbf{\textsc{LogicSeg}}} & \multirow{-2}{*}{Swin-S}   & \reshl{\textbf{94.31}}{1.35}  & \reshl{\textbf{83.85}}{1.28} \\
      \bottomrule[1.0pt]
    \end{tabular}
  }
  \vspace{-7pt}
  \captionsetup{font=small}
  \caption{\textbf{$_{\!}$Urban$_{\!}$ scene$_{\!}$ parsing$_{\!}$ results$_{\!}$} (\S\ref{sec:4.1})$_{\!}$ on$_{\!}$ Cityscapes$_{\!}$~\cite{cordts2016cityscapes} \texttt{val}~with a two-level label hierarchy of $6$/$19$ concepts.}
  \label{table:cityscape}
  \vspace{-12pt}
\end{table}

\subsection{Quantitative Comparison Result}\label{sec:4.1}
\noindent\textbf{Mapillary$_{\!\!}$ Vistas$_{\!}$ 2.0$_{\!}$~\cite{neuhold2017mapillary}$_{\!}$ \texttt{val}.}$_{\!}$  From$_{\!}$ Table$_{\!}$~\ref{table:mapillary}$_{\!}$ we$_{\!}$ can$_{\!}$ observe that our approach provides notable performance gains over the baselines. For example, our algorithm promotes classic DeepLabV3+$_{\!}$~\cite{chen2018encoder} by \textbf{3.65}\%/\textbf{3.42}\%/\textbf{3.29}\% over the three semantic levels. On top of MaskFormer$_{\!}$~\cite{cheng2021maskformer}, our algorithm further lifts the scores by \textbf{2.42}\%/\textbf{2.73}\%/\textbf{2.96}\%, suppressing previous hierarchy-agnostic models, as well as \textsc{Hssn}~\cite{li2022deep} -- a newly proposed hierarchy-aware segmentation model.

\noindent\textbf{Cityscapes$_{\!}$~\cite{cordts2016cityscapes} \texttt{val}.}   Table~\ref{table:cityscape} confirms again our compelling performance in challenging urban street scenes and wide benefits for different segmentation models, \ie,  \textbf{1.21}\%/\textbf{1.12}\% over DeepLabV3+, and \textbf{1.35}\%/\textbf{1.28}\% over MaskFormer. Though both encoding concept structures into segmentation, our algorithm greatly outperforms \textsc{Hssn}, suggesting the superiority of our logic reasoning framework.

\noindent\textbf{Pascal-Part-108$_{\!}$~\cite{chen2014detect} \texttt{test}.} As illustrated by Table~\ref{table:part108}, our algorithms yields remarkable performance on explaining the compositionality of object-centric semantic structures. Specifically, our algorithm not only consistently boosts the performance of base segmentation models$_{\!}$~\cite{chen2018encoder,cheng2022masked}, but also defeats two outstanding hierarchy-agnostic competitors~\cite{badrinarayanan2017segnet,singh2022float} as well as three structured alternatives~\cite{michieli2020gmnet,singh2022float,li2022deep}.

\begin{table}
  \centering
  \small
  \resizebox{0.85\columnwidth}{!}{
    \setlength\tabcolsep{5pt}
    \renewcommand\arraystretch{1.1}
    \begin{tabular}{rc|cc}
      \toprule[1.0pt]
      {Method} & {Backbone}  &$\text{mIoU}^2$$\uparrow$& $\text{mIoU}^1$$\uparrow$ \\
      \midrule
      SegNet~\cite{badrinarayanan2017segnet} & ResNet-101  & 59.81 & 36.42\\
      FCN-8s~\cite{long2015fully} & ResNet-101 & 62.26 & 38.62 \\
      BSANet~\cite{zhao2019multi}& ResNet-101  & 69.37 & 47.36 \\
      GMNet~\cite{michieli2020gmnet} & ResNet-101  & 69.28 & 47.21 \\
      FLOAT~\cite{singh2022float}  & ResNet-101  & 70.03 & 48.08 \\
      \textsc{Hssn}~\cite{li2022deep}&  ResNet-101  & 72.91 & 48.32 \\
      \textsc{Hssn}~\cite{li2022deep}&  Swin-S   & 77.01 & 54.79 \\
      \hline
      DeepLabV3+~\cite{chen2018encoder}&   & 70.86 & 46.54 \\
      + \textbf{\textsc{LogicSeg}} & \multirow{-2}{*}{ResNet-101}  & \reshl{\textbf{73.68}}{2.82} & \reshl{\textbf{49.13}}{2.69} \\
      \hline
      MaskFormer~\cite{cheng2021maskformer} & & 75.78 & 53.07 \\
      + \textbf{\textsc{LogicSeg}} & \multirow{-2}{*}{Swin-S}   &  \reshl{\textbf{77.92}}{2.14}&  \reshl{\textbf{55.53}}{2.46} \\
      \bottomrule[1.0pt]
    \end{tabular}
  }
  \vspace{-7pt}
  \captionsetup{font=small}
  \caption{\small\textbf{Object part parsing results} (\S\ref{sec:4.1}) on PASCAL-Part-108$_{\!}$~\cite{chen2014detect} \texttt{test} with a two-level label hierarchy of $20$/$108$ concepts.}
  \label{table:part108}
  \vspace{-8pt}
\end{table}

\begin{table}
  \centering
  \small
  \resizebox{0.99\columnwidth}{!}{
    \setlength\tabcolsep{3.5pt}
    \renewcommand\arraystretch{1.1}
    \begin{tabular}{r|c|ccc}
      \toprule[1.0pt]
      {Method} & {Backbone}&  $\text{mIoU}^3$$\uparrow$&$\text{mIoU}^2$$\uparrow$& $\text{mIoU}^1$$\uparrow$ \\
      \midrule
      OCRNet~\cite{yuan2020object}&  HRNet-W48 & 76.33& 55.76 & 44.92\\
      SETR~\cite{zheng2021rethinking} & ViT-L & 78.92 & 59.03 & 49.41 \\
      UperNet~\cite{xiao2018unified}&  Swin-S & 78.90 & 59.17 & 49.47 \\
      SegMentor~\cite{strudel2021segmenter} & ViT-S & 77.32 & 57.18 & 46.82  \\
      K-Net~\cite{zhang2021k} & Swin-S & 79.11 & 59.38 & 49.95 \\
      SegFormer~\cite{xie2021segformer}& MiT-B4 & 79.85 & 60.24 & 51.08 \\
      GMMSeg~\cite{lianggmmseg} & MiT-B5 & 80.13 & 60.91 & 52.12 \\
      Mask2Former~\cite{cheng2022masked}  & Swin-S & 80.46 & 61.15 & 52.43 \\
      \textsc{Hssn}~\cite{li2022deep}  & ResNet-101 & 79.23 & 58.52 & 47.69\\
      \textsc{Hssn}~\cite{li2022deep}  & Swin-S & 82.59 & 62.56 & 52.37\\
      \hline
      DeepLabV3+~\cite{chen2018encoder}&  & 77.24 & 56.87 & 46.43 \\
      {+ \textbf{\textsc{LogicSeg}}} & \multirow{-2}{*}{ResNet-101}  & \reshl{\textbf{79.60}}{2.36} & \reshl{\textbf{59.04}}{2.17} & \reshl{\textbf{48.46}}{2.03} \\
      \hline
      MaskFormer~\cite{cheng2021maskformer}&  & 79.89 &60.32 & 51.04 \\
      {+ \textbf{\textsc{LogicSeg}}} & \multirow{-2}{*}{Swin-S}  & \reshl{\textbf{82.45}}{2.56}  &\reshl{\textbf{62.44}}{2.12} & \reshl{\textbf{52.82}}{1.78}\\
      \bottomrule[1.0pt]
    \end{tabular}
  }
  \vspace{-7pt}
  \captionsetup{font=small}
  \caption{\small\textbf{Generic scene parsing results} (\S\ref{sec:4.1}) on ADE20K$_{\!}$~\cite{zhou2017scene} \texttt{val} with a three-level label hierarchy of $3$/$14$/$150$ concepts.}
  \label{table:ade20k}
  \vspace{-12pt}
\end{table}

\noindent\textbf{ADE20K\!~\cite{zhou2017scene} \texttt{val}.} Table$_{\!}$~\ref{table:ade20k} presents our parsing results in general scenes. With a relatively conservative baseline, \ie, DeepLabV3+$_{\!}$~\cite{chen2018encoder}, our algorithm earn \textbf{79.60}\%, \textbf{59.04}\%, and \textbf{48.46}\%, in terms of mIoU$^1$, mIoU$^2$, and mIoU$^3$ respectively. It delivers a solid overtaking against Mask2Former$_{\!}$~\cite{cheng2022masked}, which is$_{\!}$ built$_{\!}$ upon$_{\!}$ a$_{\!}$ more$_{\!}$ advanced$_{\!}$ architecture.$_{\!}$ When$_{\!}$ applied to MaskFormer$_{\!}$~\cite{cheng2021maskformer}, our algorithm achieves \textbf{82.45}\%/ \textbf{62.44}\%/\textbf{52.82}\%, pushing forward the state-of-the-art.

\begin{figure*}[t]
   \vspace{-10px}
  \begin{center}
    \includegraphics[width=\linewidth]{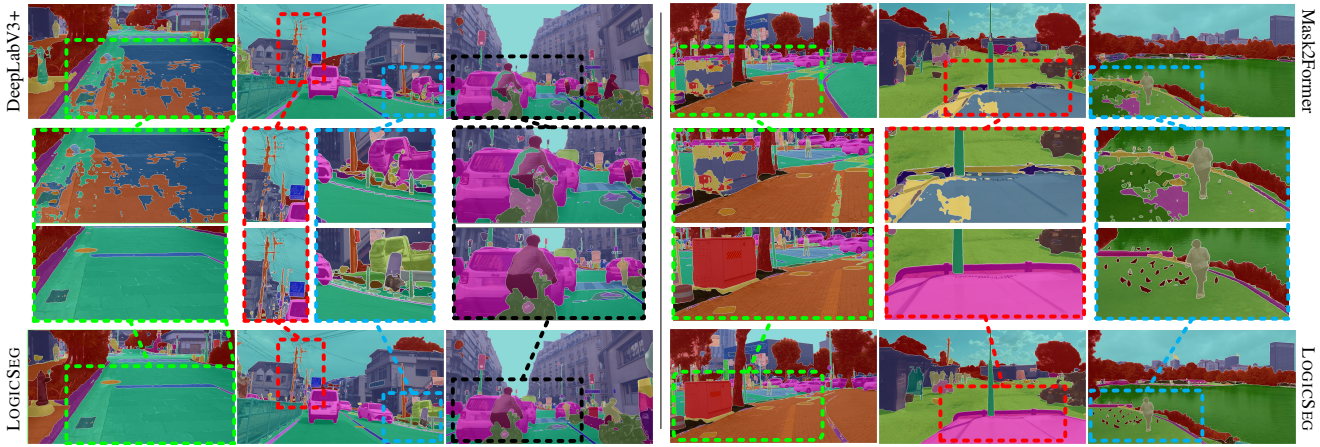}
    \put(-494.2,126){\rotatebox{90}{\scriptsize DeepLabV3+}}
    \put(-4.2,165){\rotatebox{270}{\scriptsize Mask2Former}}
    \put(-494.2,5){\rotatebox{90}{\scriptsize \textsc{LogicSeg}}}
    \put(-4.2,37){\rotatebox{270}{\scriptsize \textsc{LogicSeg}}}
  \end{center}
   \vspace{-15px}
  \caption{\small{\textbf{Visual results} (\S\ref{sec:4.2}) on$_{\!}$ Mapillary$_{\!}$ Vistas$_{\!}$ 2.0$_{\!}$~\cite{neuhold2017mapillary}. \textit{Left}: DeepLabV3+$_{\!}$~\cite{chen2018encoder} \textit{vs} \textsc{LogicSeg}; \textit{Right}: Mask2Former~$_{\!}$\cite{cheng2022masked} \textit{vs} \textsc{LogicSeg}. }}
  \label{fig:vis}
  \vspace{-5px}
\end{figure*}

\begin{table*}[t]
\centering
    \vspace{-1.em}
   \subfloat[{logic loss (\S\ref{sec:training})}\label{table:ablation1}]{%
        \resizebox{0.4265\textwidth}{!}{
        \setlength\tabcolsep{2.3pt}
      \begin{tabular}{ccc|ccc|c}
      \toprule[1.0pt]
       $\mathcal{L}_C$  & $\mathcal{L}_D$  & $\mathcal{L}_E$  & \multirow{1}{*}{$\text{mIoU}^3$$\uparrow$} &\multirow{1}{*}{$\text{mIoU}^2$$\uparrow$}& \multirow{1}{*}{$\text{mIoU}^1$$\uparrow$} & Training Speed\\
       Eq.\!~\ref{eq:CL1}  & Eq.\!~\ref{eq:DL1} & Eq.\!~\ref{eq:EL1}  & \multicolumn{3}{c|}{{\color{gray}(multi scale)}} & (min/epoch)\\
      \midrule
      & & & 81.86 & 68.17 & 37.43 & 45.62 \\
      \cmark &  &  & 83.56  & 69.74 & 38.71 & \delay{46.35}{+1.60\%}\\
      & \cmark &  & 84.08 & 69.97 & 38.98 & \delay{46.13}{+1.12\%}\\
      &  & \cmark & 83.42 & 69.60 & 38.43 &  \delay{46.72}{+2.41\%}\\
      \cmark & \cmark & \cmark & \textbf{85.51} & \textbf{71.69} & \textbf{40.72} & \delay{47.67}{+4.51\%}\\
       \bottomrule[1.0pt]
    \end{tabular}
   }
   }
   \subfloat[{iteration of message passing (\S\ref{sec:inference})}\label{table:ablation2}]{%
      \resizebox{0.3365\textwidth}{!}{
      \setlength\tabcolsep{2.5pt}
      \begin{tabular}{c|ccc|c}
      \toprule[1.0pt]
      $\#$  & $\text{mIoU}^3$$\uparrow$ &$\text{mIoU}^2$$\uparrow$& $\text{mIoU}^1$$\uparrow$ & Inference Speed\\
       Iter.  & \multicolumn{3}{c|}{{\color{gray}(multi scale)}} & (fps)\\
      \midrule
      0 & 84.62 &70.95  & 40.18 & 3.44 \\
      1 & 85.23 & 71.46 & 40.56 &  \delay{3.37}{-2.03\%} \\
      \textbf{2} & \textbf{85.51} & \textbf{71.69} & \textbf{40.72} & \delay{3.31}{-3.78\%} \\
      3 & 84.84 &  71.12& 40.29 & \delay{3.25}{-5.52\%}\\
      4 & 84.49 & 70.84 & 40.03 &  \delay{3.20}{-6.98\%}\\
      \bottomrule[1.0pt]
    \end{tabular}
   }
   }
   \subfloat[{aggregation coefficient for $\forall$ (Eq.$_{\!\!}$~\ref{eq:fq})}\label{table:ablation4}]{
   \resizebox{0.23\textwidth}{!}{
   \setlength\tabcolsep{3.5pt}
   \begin{tabular}{c|ccc}
      \toprule[1.0pt]
       \multirow{2}{*}{$q$}&$\text{mIoU}^3$$\uparrow$ &$\text{mIoU}^2$$\uparrow$& $\text{mIoU}^1$$\uparrow$ \\
       &\multicolumn{3}{c}{{\color{gray}(multi scale)}}\\
       \midrule
       1 & 83.83 & 70.22 & 38.77 \\
       3 & 84.65 & 71.15 & 40.09 \\
         \textbf{5} &  \textbf{85.51} & \textbf{71.69} & \textbf{40.72}  \\
       8 & 84.47 & 71.03 & 39.74 \\
       10& 83.52 & 69.88 & 38.25 \\
      \bottomrule[1.0pt]
   \end{tabular}
   }
   }

   \vspace{-7pt}
   \captionsetup{font=small}
   \caption{{\textbf{Ablative$_{\!}$ studies}$_{\!}$ on$_{\!}$ Mapillary$_{\!}$ Vistas$_{\!}$ 2.0$_{\!}$~\cite{neuhold2017mapillary}$_{\!}$ \texttt{val}$_{\!}$ (\S\ref{sec:ablation}).$_{\!}$ All$_{\!}$ variants$_{\!}$ are$_{\!}$ based$_{\!}$ on$_{\!}$ DeepLabV3+$_{\!}$~\cite{chen2018encoder}$_{\!}$ with$_{\!}$ ResNet-101$_{\!}$~\cite{he2016deep}$_{\!}$ backbone.}}
   \label{ablative:merge}
     \vspace{-14pt}
\end{table*}

Taking together, our extensive benchmarking results provide solid evidence that our algorithm successfully unlocks the power of logic reasoning in large-scale visual parsing, and owns broad applicability across various task scenarios, segmentation architectures, and backbone networks.

\subsection{Qualitative Comparison Result}\label{sec:4.2}
Fig.\!~\ref{fig:vis} visualizes qualitative comparisons of \textsc{LogicSeg} against DeepLabV3+\!~\cite{chen2018encoder} (\textit{left}) and Mask2Former~\cite{cheng2022masked} (\textit{right}) on Mapillary Vistas 2.0 dataset\!~\cite{neuhold2017mapillary}. As seen, with the help of symbolic reasoning, \textsc{LogicSeg} can generate higher-quality predictions even in challenging scenarios.

\subsection{Diagnostic Experiment}\label{sec:ablation}

For thorough evaluation, we perform a series of ablative studies on Mapillary Vistas 2.0$_{\!}$~\cite{neuhold2017mapillary} \texttt{val}. All variants are based on DeepLabV3+$_{\!}$~\cite{chen2018encoder} with ResNet-101~\cite{he2016deep} backbone.

\noindent\textbf{Logic-Induced Training.} We first study the effectiveness of our logic-induced training strategy (\S\ref{sec:training}) in Table~\ref{table:ablation1}. 1$^{st}$ row reports the results of our baseline model -- DeepLabV3+.  2$^{nd}$, 3$^{rd}$, and 4$^{th}$ rows respectively list the scores obtained by individually adding our \textit{C}-rule$_{\!}$ loss $\mathcal{L}_C$ (\textit{cf}.$_{\!}$~Eq.$_{\!\!}$~\ref{eq:CL2}), \textit{D}-rule$_{\!}$ loss $\mathcal{L}_D$ (\textit{cf}.$_{\!}$~Eq.$_{\!\!}$~\ref{eq:DL2}), and \textit{E}-rule$_{\!}$ loss $\mathcal{L}_E$ (\textit{cf}.$_{\!}$~Eq.$_{\!\!}$~\ref{eq:EL2}). The last row gives the performance of our full loss $\mathcal{L}$ (\textit{cf}.$_{\!}$~Eq.$_{\!\!}$~\ref{eq:AL}). We can find that: \textbf{i)} Taking each of our logic losses into consideration can provide consistent performance gains. This demonstrates that different logic rules can describe different properties of semantic structure and verify that the segmentation model can indeed benefit from our proposed logic losses. \textbf{ii)} Combing all three logic losses together results in the best performance. This suggests that our logic rules provide a comprehensive description of the relational knowledge in the semantic hierarchy $\mathcal{T}$, and supports our core idea that exploiting symbolic knowledge is crucial for visual semantic interpretation and can boost sub-symbolic learning.

\noindent\textbf{Training$_{\!}$ Speed.$_{\!}$} As$_{\!}$ shown$_{\!}$ in$_{\!}$ the$_{\!}$ last$_{\!}$ column$_{\!}$ of$_{\!}$ Table$_{\!}$~\ref{table:ablation1}, our logic-induced$_{\!}$ training$_{\!}$ regime$_{\!}$ causes$_{\!}$ a trivial$_{\!}$ delay$_{\!}$ ($\sim$5.0\%).

\noindent\textbf{Logic-Induced Inference.} We next investigate the impact of our logic-induced inference strategy (\S\ref{sec:inference}) in Table~\ref{table:ablation2}. 1$^\text{st}$ row reports the results of network feed-forward output. The rest rows give the scores obtained with different iterations of message passing (\textit{cf}.$_{\!}$~Eq.$_{\!}$~\ref{eq:iu}). These results demonstrate the efficacy of our strategy and the necessity of incorporating logic reasoning into network inference. We accordingly set 2-iteration as the default to pursue the best performance.

\noindent\textbf{Inference$_{\!}$ Speed.$_{\!}$} We also report inference speed (fps) in Table~\ref{table:ablation2}. As seen, our logic-induced inference strategy only slows the speed slightly during model deployment ($\sim$3.8\%).

\noindent\textbf{Aggregation Coefficient.} For the approximation of $\forall$ quantifier (\textit{cf}.$_{\!}$~Eq.$_{\!\!}$~\ref{eq:fq}), we adopt the generalized mean for stable training,$_{\!}$ as$_{\!}$ suggested$_{\!}$ by$_{\!}$~\cite{badreddine2022logic}.$_{\!}$ Basically,$_{\!}$ a$_{\!}$ higher$_{\!}$ coefficient$_{\!}$ $q$ renders $\forall$ a stronger focus on outliers. For completeness, the results with different values of $q$ are reported in Table~\ref{table:ablation4}. 

\section{Conclusion and Discussion}
The creation of intelligent systems that integrate the fundamental cognitive abilities of reasoning and learning has long been viewed as a core challenge for AI~\cite{valiant2003three}. While the community recently witnessed great advances in high-level perception tasks such as visual semantic interpretation, top-leading solutions are purely driven by sub-symbolic learning, far from such effective integration. The present study repre- sents$_{\!}$  an$_{\!}$  innovative$_{\!}$  and$_{\!}$  solid$_{\!}$  attempt$_{\!}$  towards$_{\!}$  closing$_{\!}$  this$_{\!}$  gap. By embedding symbolic logic into both network training and inference, a structured and powerful visual semantic parser is delivered. We hope this work can stimulate our community to rethink current \textit{de facto}, sub-symbolic paradigm and inves- tigate new methodologies,$_{\!}$ from the perspective of achieving a better understanding of human and machine intelligence.

{\noindent\small{\textbf{Acknowledgements}  This work was supported in part by the Australian Research Council (ARC) under Grant DP200100938.

{\small
\bibliographystyle{ieee_fullname}
\bibliography{egbib}
}

\clearpage
\appendix

In this document, we first provide the pseudo code of \textsc{LogicSeg}
 in \S\ref{sec:code}. We next show the detailed label hierarchy for each dataset in \S\ref{sec:hiera}. 
In addition, we offer more qualitative results in \S\ref{sec:vis}. Finally, we discuss the limitations and border impact of our algorithm in \S\ref{sec:lim} and \S\ref{sec:impact}, respectively. To ensure reproducibility and foster future research, our full implementation will be released after acceptance.

\section{Pseudo Code} \label{sec:code}
To facilitate a comprehensive understanding of \textsc{LogicSeg}, we provide pseudo code for the logic-induced inference (\S{\color{red}{3}}) of \textsc{LogicSeg} in Algorithm \ref{alg:inference} and Algorithm \ref{alg:score}, respectively. It can be seen that all the message creation processes are implemented in matrix operation which can enjoy the acceleration of the parallel architecture of GPUs. The \textit{for-loop} is merely adopted in Algorithm \ref{alg:inference} when normalizing or summarizing the prediction in a level-wise manner, with $O(n)$ time complexity.

\section{Label Hierarchy} \label{sec:hiera}
For Mapillary Vistas 2.0\!~\cite{neuhold2017mapillary} and Cityscapes\!~\cite{cordts2016cityscapes} datasets, we adopt the officially provided label hierarchies following\!~\cite{li2022deep}. For Pascal-Part-108\!~\cite{chen2014detect}, we use the hierarchy defined in \cite{michieli2020gmnet,singh2022float}. For ADE20K\!~\cite{zhou2017scene}, we organize a three-level label hierarchy by considering the semantic relations between labels according to the WordNet\!~\cite{miller1995wordnet}. The detailed label hierarchies for each datasets are provided in Fig.\!~\ref{fig:map_hiera}-\ref{fig:ade_hiera}.

\section{More Qualitative Comparison Result}\label{sec:vis}
We  provide more visual results that compare \textsc{LogicSeg} to Mask2Former\!~\cite{cheng2022masked} and to DeeplabV3+\!~\cite{chen2018encoder} in Fig.\!~\ref{fig:map_sm}-\ref{fig:city_sm} and Fig.\!~\ref{fig:ade_sm}-\ref{fig:part_sm}, respectively. It can be observed that \textsc{LogicSeg} performs robust in hard cases and can consistently deliver more satisfying results compared with the baseline algorithms.

\vspace{1pt}
\section{Limitation} \label{sec:lim}
Currently our algorithm is specifically designed for tree-shape label hierarchy. It is interesting to extend our algorithm to handle more complicated and real-world semantic structures, for example, parent classes sharing some child classes. We leave this as a part of our future work.

\vspace{1pt}
\section{Border Impact} \label{sec:impact}
This paper contributes to research on intelligent scene understanding, and thus is expected to eventually benefit automatic driving, education, health care, and economic development of the human society. Moreover, although our algorithm is able to parse the hierarchical relations between semantic concepts and yields improved performance over current top-leading competitors, the relevant security measures still need to be erected and caution should always be exercised.

\begin{algorithm}[bp]
\renewcommand\thealgorithm{S1}
\caption{Pseudo-code for logic-induced inference of \textsc{LogicSeg} in a PyTorch-like style (Part I).}
\label{alg:inference}
\definecolor{codeblue}{rgb}{0.25,0.5,0.5}
\lstset{
  backgroundcolor=\color{white},
  basicstyle=\fontsize{7.2pt}{7.2pt}\ttfamily\selectfont,
  columns=fullflexible,
  breaklines=true,
  captionpos=b,
  escapeinside={(:}{:)},
  commentstyle=\fontsize{7.2pt}{7.2pt}\color{codeblue},
  keywordstyle=\fontsize{7.2pt}{7.2pt},
}
\begin{lstlisting}[language=python]
"""
T: index matrix indicates the hierarchy, for example:
              a     (:\ \ \ \ \ \ \ \ \ \ \ \ \ \ \!\!\!\!:)  a b c d e f
             / \        (:\ \ \ \ \ \ \ \ \ \ \!:)    a 0 1 1 0 0 0
            b  (:\ \!\!:)  c  (:\ \ :)   -->   (:\ \ :)   b 0 0 0 1 1 0
           /\   (:\ :) \ (:\ \ \ \ \ \ \ \ \ \!\!\!:)    c 0 0 0 0 0 1
          d (:\ \!\!:)e (:\ \ \!\!\!\!:) f
P: matrix indicates the peer relation, for example:
                                    (:\!\!:)a b c d e f
              a     (:\ \ \ \ \ \ \ \ \ \ \ \ \!\!\!\!:)  a 0 0 0 0 0 0
             / \        (:\ \ \ \ \ \ \ \ \ \ \!:)    b 0 0 1 0 0 0
            b  (:\ \!\!:)  c  (:\ \ :)   -->   (:\ \ :)   c 0 1 0 0 0 0
           /\   (:\ :) \ (:\ \ \ \ \ \ \ \ \ \!\!\!:)    d 0 0 0 0 1 1
          d (:\ \!\!:)e (:\ \ \!\!\!\!:) f (:\ \ \ \ \ \ \ \ $_{\!}$\!:)      e 0 0 0 1 0 1
                                 (:\!:)f 0 0 0 1 1 0
V: array stores the class number in each
    hierarchical level, V[l] = (:\color{codegreen}{$|\mathcal{V}^{l+1}|$}:)
R: round of message passing
L: number of hierarchical level
s_k: grounded predicates ((:\color{codegreen}{$|\mathcal{V}|$}:) x HW)
"""

(:\color{codedefine}{\textbf{def}}:) (:\color{codefunc}{\textbf{message\_passing}}:)(s_k):
    s_k += (:\color{codecall}{\textbf{c\_score}}:)(s_k)  + (:\color{codecall}{\textbf{d\_score}}:)(s_k) + (:\color{codecall}{\textbf{e\_score}}:)(s_k)
    # hierarchical level-wise normalization
    n = 0
    (:\color{codedefine}{\textbf{for}}:) l (:\color{codedefine}{\textbf{in}}:) (:\color{codedefine}{range}:)(L, 0, -1):
        s_k[n:n+V[l]] = s_k[n:n+V[l]].(:\color{codepro}{\textbf{softmax}}:)((:\color{codedim}{\textbf{dim}}:)=0)
        n += V[l]

    (:\color{codedefine}{\textbf{return}}:) s_k

(:\color{codedefine}{\textbf{def}}:) (:\color{codefunc}{\textbf{inference}}:)(s_k):
    # R times of message passing
    (:\color{codedefine}{\textbf{for}}:) t (:\color{codedefine}{\textbf{in}}:) (:\color{codedefine}{range}:)(R):
        s_k = (:\color{codecall}{\textbf{message\_passing}}:)(s_k)

    # (N_p x (:\color{codegreen}{$|\mathcal{V}|$}:) x 1)*(1 x (:\color{codegreen}{$|\mathcal{V}|$}:) x HW)
    s_f = T.(:\color{codepro}{\textbf{unsqueeze}}:)(-1) * s_k.(:\color{codepro}{\textbf{unsqueeze}}:)(0)
    n = V[L-1]
    t_s = s_f[:V[L-1]]
    #---------top-scoring path (Eq. 18)----------#
    (:\color{codedefine}{\textbf{for}}:) l (:\color{codedefine}{\textbf{in}}:) (:\color{codedefine}{range}:)(L-2, -1, -1):
        t_s = t_s.(:\color{codepro}{\textbf{unsqueeze}}:)(1)
        # ((:\color{codegreen}{$|\mathcal{V}^l|$}:) x (:\color{codegreen}{$|\mathcal{V}|$}:) x HW) + ((:\color{codegreen}{$|\mathcal{V}^l|$}:) x 1 x HW)
        t_s = (s_f[n:n+V[l]] + t_s)
        # ((:\color{codegreen}{$|\mathcal{V}^l|$}:) x (:\color{codegreen}{$|\mathcal{V}|$}:) x HW)  --> ((:\color{codegreen}{$|\mathcal{V}|$}:) x HW)
        t_s = t_s*(T[n:n+V[l]].unsqeeze(-1)).(:\color{codepro}{\textbf{sum}}:)(0)
        n += V[l]
        # ((:\color{codegreen}{$|\mathcal{V}^{l-1}|$}:) x HW)
        t_s = t_s[n:n+V[l-1]]
    # ((:\color{codegreen}{$|\mathcal{V}^1|$}:) x HW) --> (HW)
    pred = t_s.(:\color{codepro}{\textbf{argmax}}:)((:\color{codedim}{\textbf{dim}}:)=0)

    (:\color{codedefine}{\textbf{return}}:) pred

\end{lstlisting}
\end{algorithm}

\begin{algorithm}[bp]
\renewcommand\thealgorithm{S2}
\caption{Pseudo-code for logic-induced inference of \textsc{LogicSeg} in a PyTorch-like style (Part II).}
\label{alg:score}
\definecolor{codeblue}{rgb}{0.25,0.5,0.5}
\lstset{
  backgroundcolor=\color{white},
  basicstyle=\fontsize{7.2pt}{7.2pt}\ttfamily\selectfont,
  columns=fullflexible,
  breaklines=true,
  captionpos=b,
  escapeinside={(:}{:)},
  commentstyle=\fontsize{7.2pt}{7.2pt}\color{codeblue},
  keywordstyle=\fontsize{7.2pt}{7.2pt},
}
\begin{lstlisting}[language=python]
"""
N_p: class number of non-leaf nodes
s_k: grounded predicates ((:\color{codegreen}{$|\mathcal{V}|$}:) x HW)
"""
(:\color{codedefine}{\textbf{def}}:) (:\color{codefunc}{\textbf{c\_score}}:)(s_k):
    #-------------C-message (Eq. 16)-------------#
    # (N_p x (:\color{codegreen}{$|\mathcal{V}|$}:) x 1)*(1 x (:\color{codegreen}{$|\mathcal{V}|$}:) x HW)
    c_f = T.(:\color{codepro}{\textbf{unsqueeze}}:)(-1) * s_k.(:\color{codepro}{\textbf{unsqueeze}}:)(0)
    # (N_p x 1 x HW)*(N_p x (:\color{codegreen}{$|\mathcal{V}|$}:) x HW)
    c_m = s_k[:N_p].(:\color{codepro}{\textbf{unsqueeze}}:)(1) * c_f
    # (:\color{codegreen}{$1\!-\!\bm{s}_k[v]\!+\! \bm{s}_k[v]\!\cdot_{\!}\bm{s}_k[p_{v}]$}:)
    c_m = 1 - c_f + c_m

    #-----gather received C-messages (Eq. 17)----#
    # (N_p x HW)
    c_s = (c_f * c_m).(:\color{codepro}{\textbf{sum}}:)((:\color{codedim}{\textbf{dim}}:)=1)
    c_s = c_s / T.(:\color{codepro}{\textbf{sum}}:)((:\color{codedim}{\textbf{dim}}:)=1).(:\color{codepro}{\textbf{unsqueeze}}:)(-1)
    # ((:\color{codegreen}{$|\mathcal{V}|$}:) x HW)
    c = torch.(:\color{codepro}{\textbf{zeros}}:)((:$|\mathcal{V}|$:), HW)
    c[:N_p, :] = c_s

    (:\color{codedefine}{\textbf{return}}:) c

(:\color{codedefine}{\textbf{def}}:) (:\color{codefunc}{\textbf{d\_score}}:)(s_k):
    #-------------D-message (Eq. 16)-------------#
    # (N_p x (:\color{codegreen}{$|\mathcal{V}|$}:) x 1)*(1 x (:\color{codegreen}{$|\mathcal{V}|$}:) x HW)
    d_f = T.(:\color{codepro}{\textbf{unsqueeze}}:)(-1) * s_k.(:\color{codepro}{\textbf{unsqueeze}}:)(0)
    # (N_p x HW)*(N_p x HW)
    d_m = s_k[:N_p] * d_f.(:\color{codepro}{\textbf{max}}:)((:\color{codedim}{\textbf{dim}}:)=1)
    # (:\color{codegreen}{$1\!-\!\bm{s}_k[v]\!+\! \bm{s}_k[v]\!\cdot\!\texttt{max}(\{\bm{s}_k[c^n_v]\}_n)$}:)
    d_m = 1-s_k[:N_p] + d_m

    #-----gather received D-messages (Eq. 17)----#
    # (N_p x HW)x(N_p x HW)
    d_s = s_k[:N_p] * d_m
    # (N_p x 1 x HW)*(N_p x (:\color{codegreen}{$|\mathcal{V}|$}:) x 1)
    d_s = d_s.(:\color{codepro}{\textbf{unsqueeze}}:)(1) * T.(:\color{codepro}{\textbf{unsqueeze}}:)(-1)
    # ((:\color{codegreen}{$|\mathcal{V}|$}:) x HW)
    d_s = d_s.(:\color{codepro}{\textbf{sum}}:)((:\color{codedim}{\textbf{dim}}:)=0)

    (:\color{codedefine}{\textbf{return}}:) d_s

(:\color{codedefine}{\textbf{def}}:) (:\color{codefunc}{\textbf{e\_score}}:)(s_k):
    #-------------E-message (Eq. 16)-------------#
    # ((:\color{codegreen}{$|\mathcal{V}|$}:) x (:\color{codegreen}{$|\mathcal{V}|$}:) x 1)*(1 x (:\color{codegreen}{$|\mathcal{V}|$}:) x HW)
    e_f = P.(:\color{codepro}{\textbf{unsqueeze}}:)(-1) * s_k.(:\color{codepro}{\textbf{unsqueeze}}:)(0)
    # ((:\color{codegreen}{$|\mathcal{V}|$}:) x 1 x HW)*((:\color{codegreen}{$|\mathcal{V}|$}:) x (:\color{codegreen}{$|\mathcal{V}|$}:) x HW)
    e_m = s_k.(:\color{codepro}{\textbf{unsqueeze}}:)(1) * e_f
    # (:\color{codegreen}{$-_{\!}~(1\!-\!\textstyle\frac{1}{M}_{\!}\!\sum_{m=1}^M\!\bm{s}_k[v]_{\!}\cdot_{\!}\bm{s}_k[a_{v}^m])$}:)
    e_m = -1+e_m.(:\color{codepro}{\textbf{sum}}:)((:\color{codedim}{\textbf{dim}}:)=1)/P.(:\color{codepro}{\textbf{sum}}:)((:\color{codedim}{\textbf{dim}}:)=1).(:\color{codepro}{\textbf{unsqueeze}}:)(-1)

    #-----gather received E-messages (Eq. 17)----#
    # ((:\color{codegreen}{$|\mathcal{V}|$}:) x HW)
    e_s = e_f.(:\color{codepro}{\textbf{sum}}:)((:\color{codedim}{\textbf{dim}}:)=1) / P.(:\color{codepro}{\textbf{sum}}:)((:\color{codedim}{\textbf{dim}}:)=1).(:\color{codepro}{\textbf{unsqueeze}}:)(-1)
    # E-message should be same for all nodes in the same hierarchical level
    e_s = e_m * e_s

    (:\color{codedefine}{\textbf{return}}:) e_s
\end{lstlisting}
\end{algorithm}

\clearpage
\begin{figure*}[t]
\renewcommand\thefigure{S1}
\centering
      \includegraphics[width=0.99\linewidth]{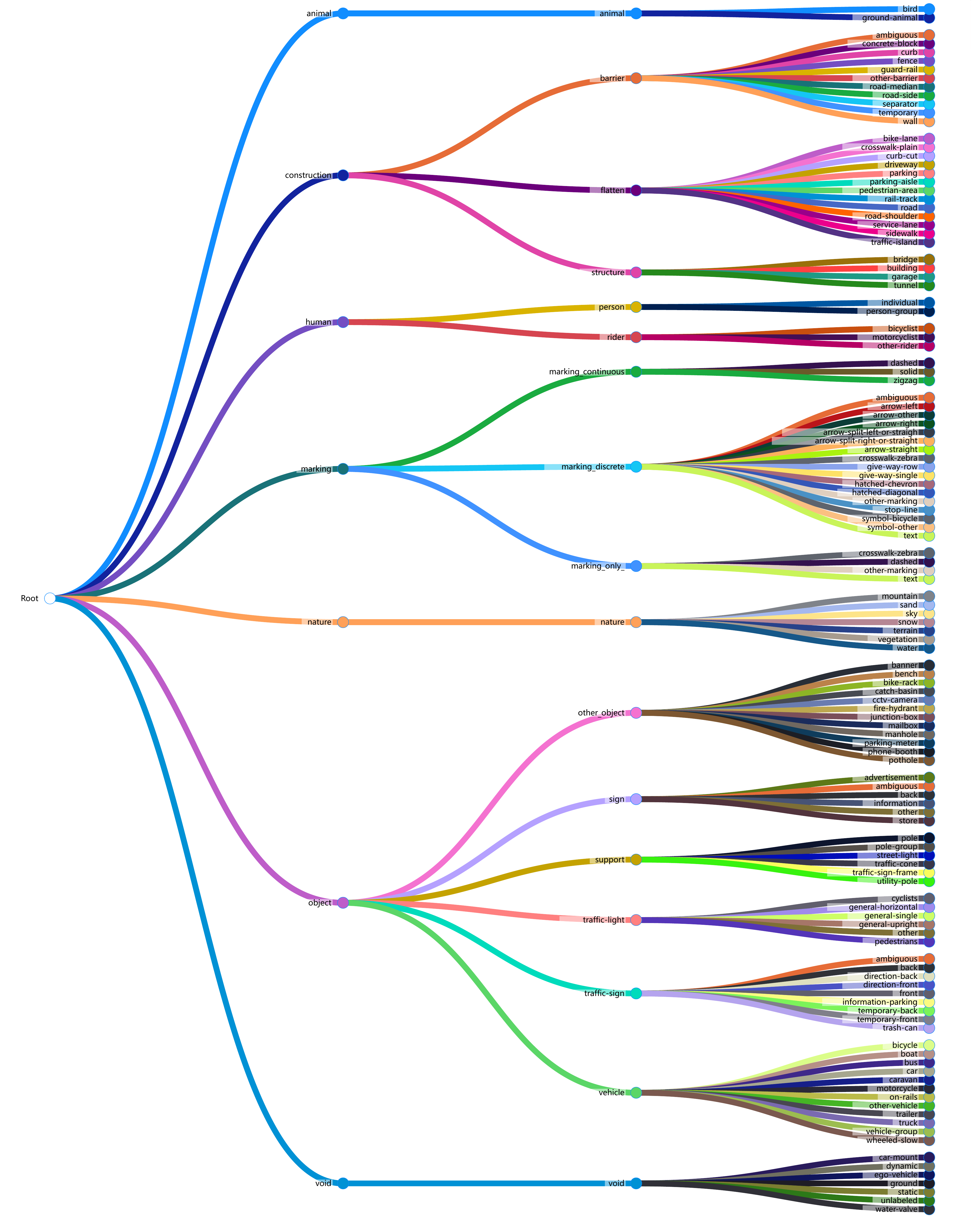}
\vspace{-3pt}
\caption{\textbf{Hierarchical label structure} of Mapillary Vistas 2.0\!~\cite{neuhold2017mapillary}.}
\label{fig:map_hiera}
\vspace{-8pt}
\end{figure*}

\begin{figure*}[t]
\renewcommand\thefigure{S2}
\centering
      \includegraphics[width=0.99\linewidth]{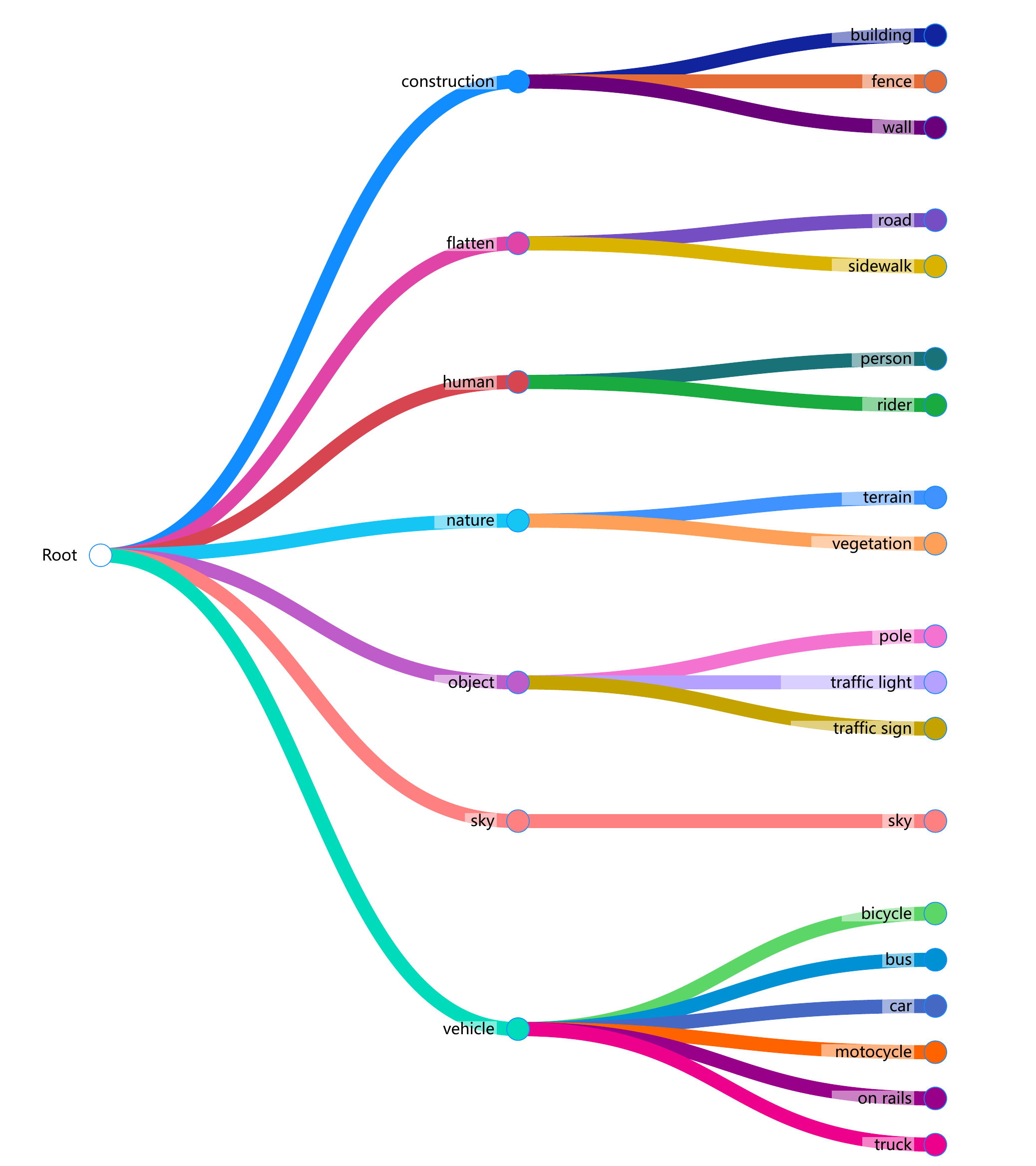}
\vspace{-3pt}
\caption{\textbf{Hierarchical label structure} of Cityscapes\!~\cite{cordts2016cityscapes}.}
\label{fig:city_hiera}
\vspace{-8pt}
\end{figure*}

\begin{figure*}[t]
\renewcommand\thefigure{S3}
\centering
      \includegraphics[width=0.99\linewidth]{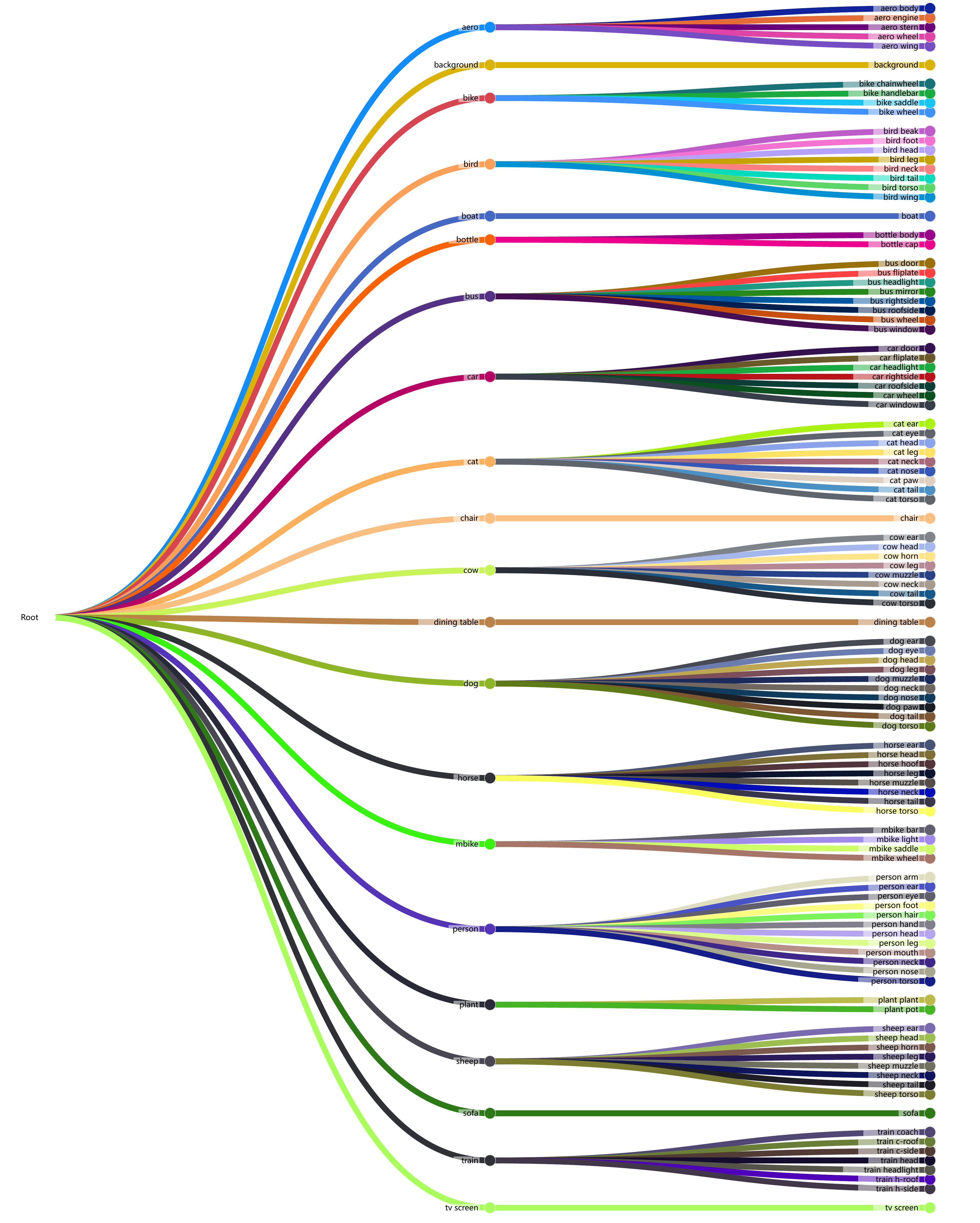}
\vspace{-3pt}
\caption{\textbf{Hierarchical label structure} of Pascal-Part-108\!~\cite{chen2014detect}.}
\label{fig:pascal_hiera}
\vspace{-8pt}
\end{figure*}

\begin{figure*}[t]
\renewcommand\thefigure{S4}
\centering
      \includegraphics[width=0.99\linewidth]{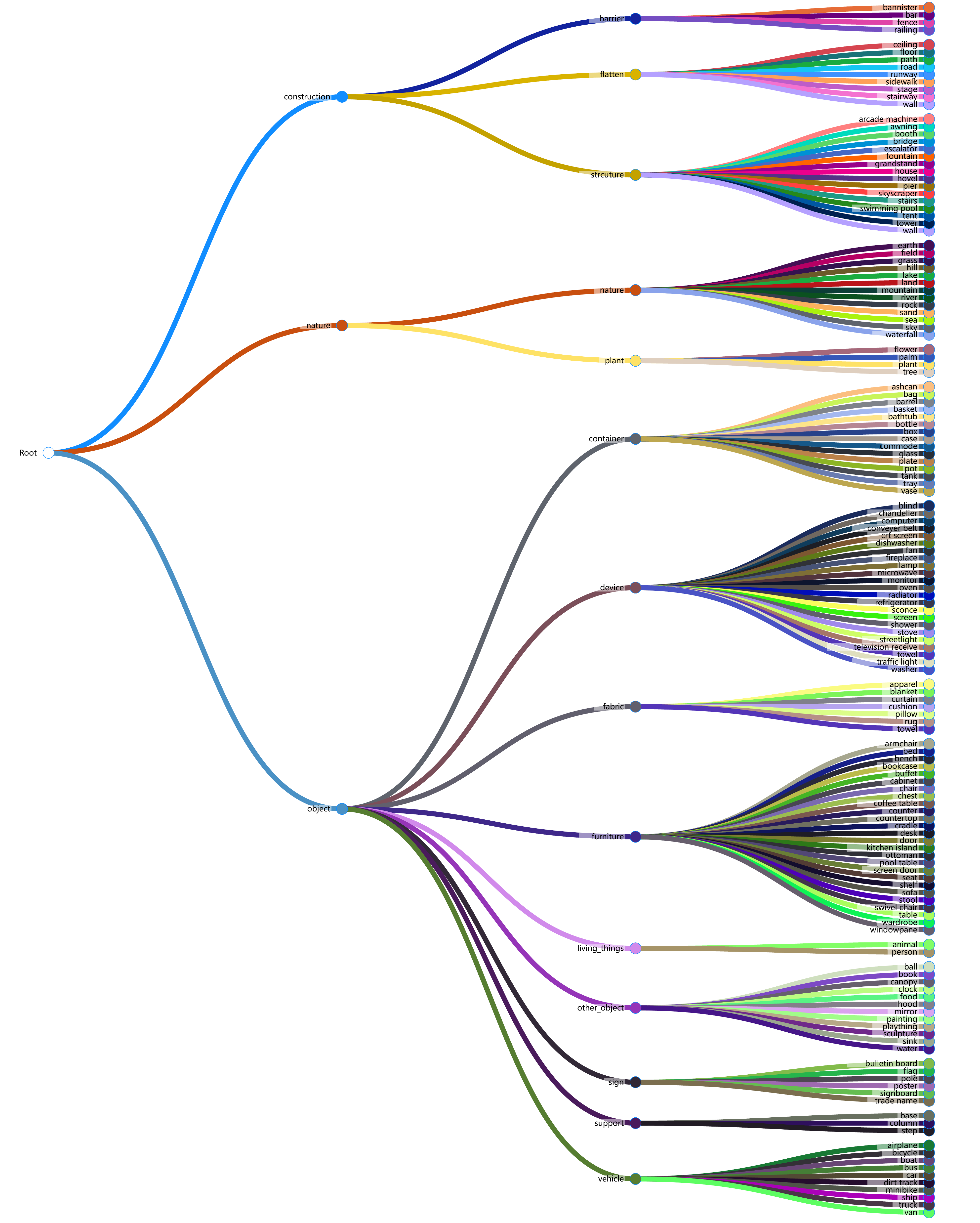}
\vspace{-3pt}
\caption{\textbf{Hierarchical label structure} of ADE20K\!~\cite{zhou2017scene}.}
\label{fig:ade_hiera}
\vspace{-8pt}
\end{figure*}

\clearpage
\begin{figure*}[t]
\renewcommand\thefigure{S5}
\centering
      \includegraphics[width=0.99\linewidth]{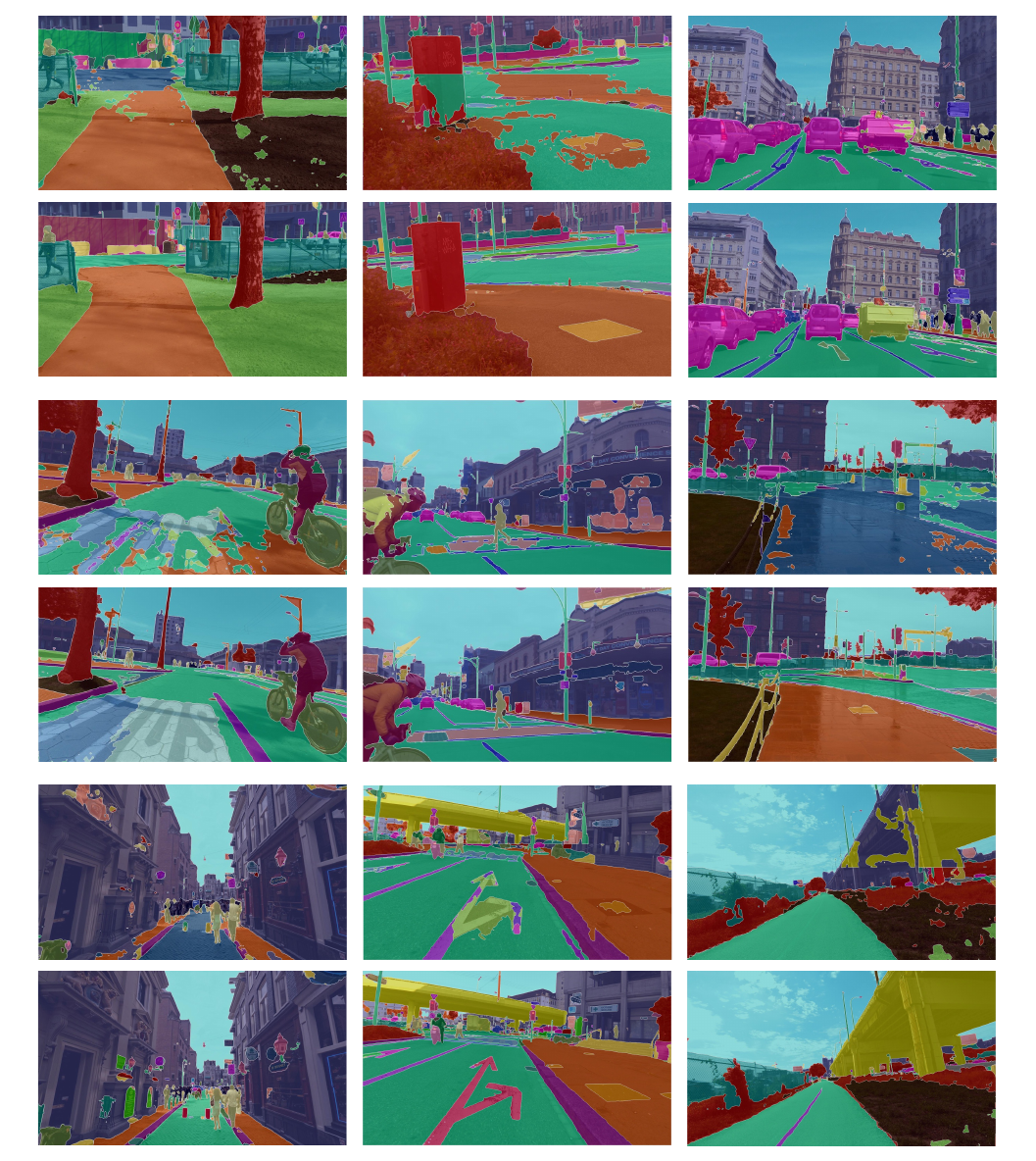}
\vspace{-3pt}
\caption{\textbf{Visual comparison results} on Mapillary Vistas 2.0\!~\cite{neuhold2017mapillary} \texttt{val}. \textit{Top}: Mask2Fomer\!~\cite{cheng2022masked} \vs \textit{Bottom}: \textsc{LogicSeg} }
\label{fig:map_sm}
\vspace{-8pt}
\end{figure*}

\begin{figure*}[t]
\renewcommand\thefigure{S6}
\centering
      \includegraphics[width=0.99\linewidth]{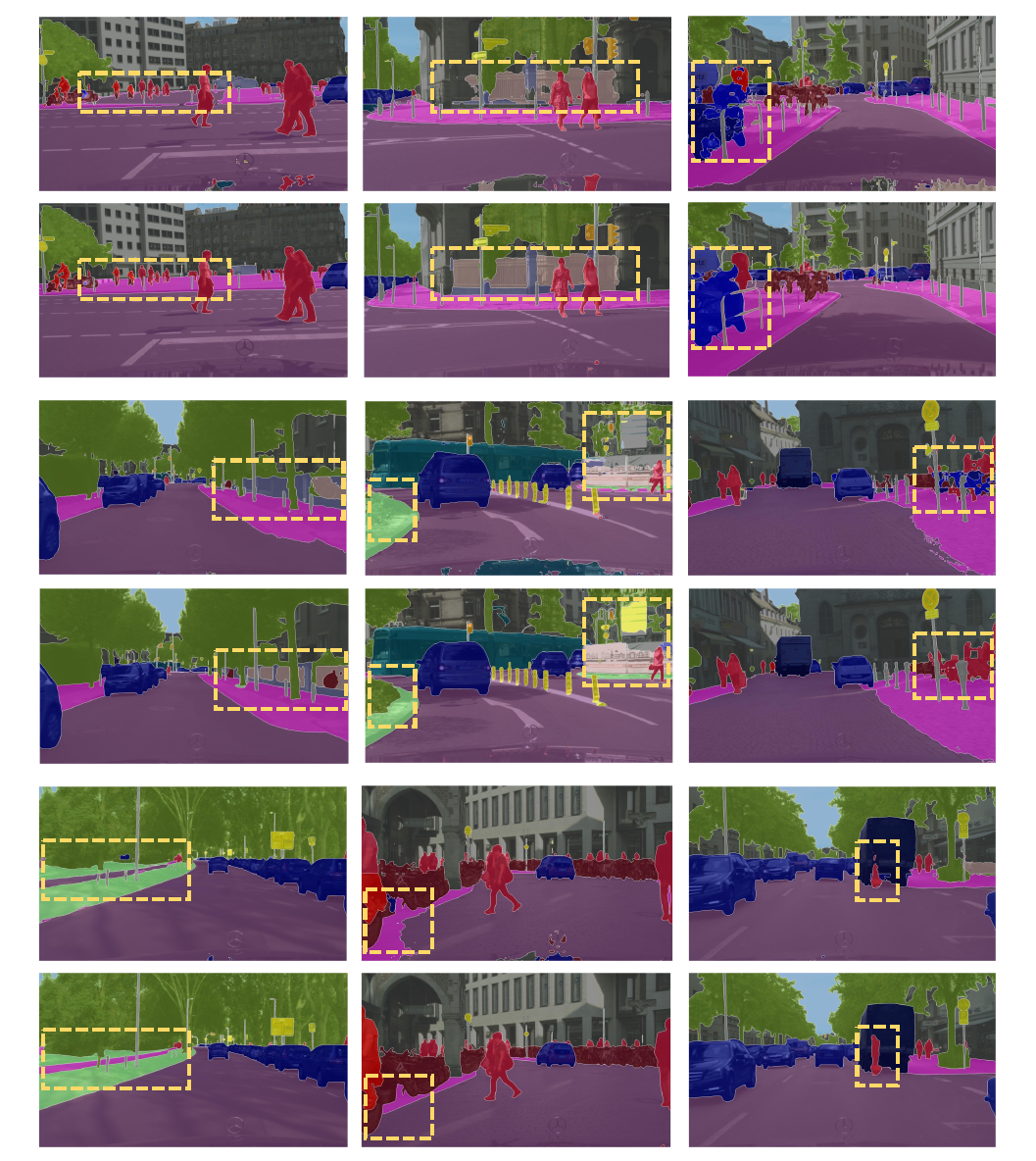}
\vspace{-3pt}
\caption{\textbf{Visual comparison results} on Cityscapes\!~\cite{cordts2016cityscapes} \texttt{val}. \textit{Top}: Mask2Fomer\!~\cite{cheng2022masked} \vs \textit{Bottom}: \textsc{LogicSeg} }
\label{fig:city_sm}
\vspace{-8pt}
\end{figure*}

\begin{figure*}[t]
\renewcommand\thefigure{S7}
\centering
      \includegraphics[width=0.99\linewidth]{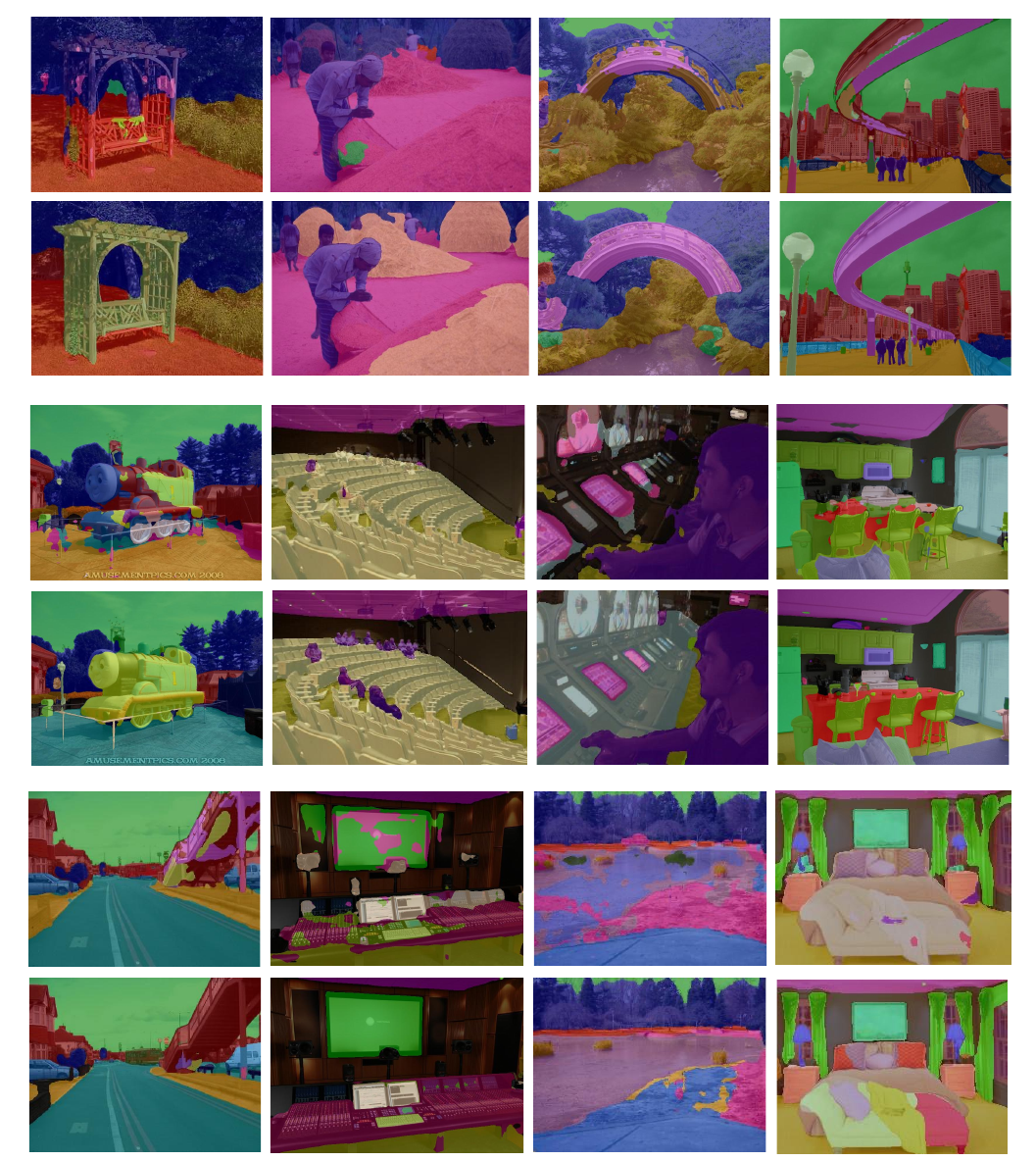}
\vspace{-3pt}
\caption{\textbf{Visual comparison results} on ADE20K\!~\cite{zhou2017scene} \texttt{val}. \textit{Top}: DeepLabV3+\!~\cite{chen2018encoder} \vs \textit{Bottom}: \textsc{LogicSeg} }
\label{fig:ade_sm}
\vspace{-8pt}
\end{figure*}

\begin{figure*}[t]
\renewcommand\thefigure{S8}
\centering
      \includegraphics[width=0.99\linewidth]{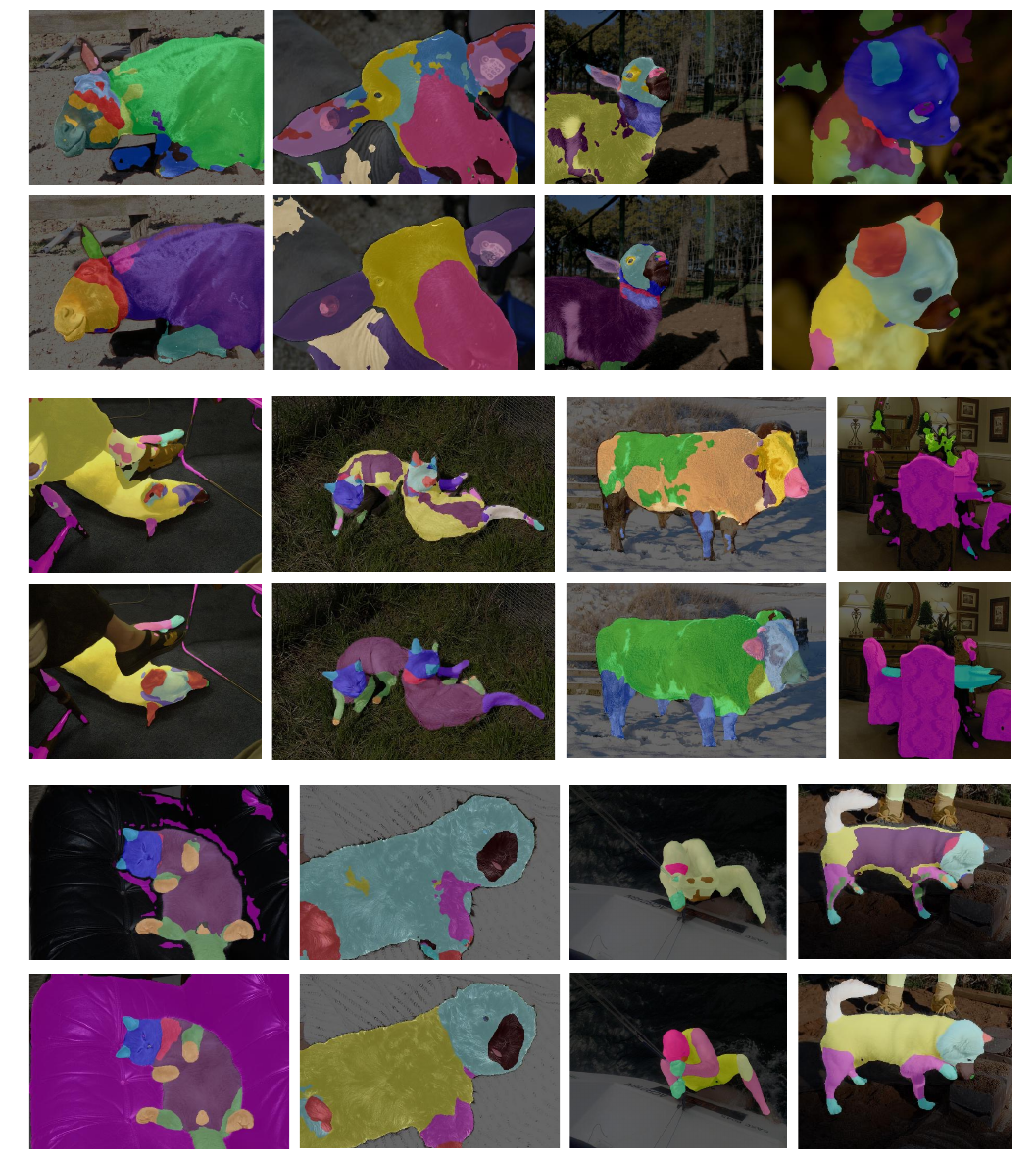}
\vspace{-3pt}
\caption{\textbf{Visual comparison results} on Pascal-Part-108\!~\cite{chen2014detect} \texttt{test}. \textit{Top}: DeepLabV3+\!~\cite{chen2018encoder} \vs \textit{Bottom}: \textsc{LogicSeg}}
\label{fig:part_sm}
\vspace{-8pt}
\end{figure*}

\end{document}